\documentclass{article}

\usepackage[final,main]{neurips_2026}

\usepackage[utf8]{inputenc}
\usepackage[T1]{fontenc}
\usepackage{hyperref}
\usepackage{url}
\usepackage{xurl}
\usepackage{booktabs}
\usepackage{amsfonts}
\usepackage{amsmath}
\usepackage{amssymb}
\usepackage{mathtools}
\usepackage{amsthm}
\usepackage{nicefrac}
\usepackage{microtype}
\usepackage{xcolor}
\usepackage{graphicx}
\usepackage{subcaption}
\usepackage{multirow}
\usepackage{array}
\usepackage{tcolorbox}
\usepackage[frozencache,cachedir=minted-cache]{minted2}
\usepackage{listings}
\usepackage{comment}
\usepackage[nameinlink]{cleveref}
\usepackage[textsize=tiny]{todonotes}

\definecolor{borderblue}{RGB}{34,51,103}
\definecolor{bggray}{RGB}{245,247,250}

\theoremstyle{plain}

\theoremstyle{definition}

\theoremstyle{remark}

\title{Program-as-Weights:\\A Programming Paradigm for Fuzzy Functions}

\author{%
  \textbf{Wentao Zhang$^{1,*}$\quad Liliana Hotsko$^{1,*}$\quad Woojeong Kim$^{2,*}$} \\[3pt]
  \textbf{Pengyu Nie$^{1}$\quad Stuart Shieber$^{3}$\quad Yuntian Deng$^{1}$} \\[5pt]
  \normalfont $^{1}$University of Waterloo\qquad $^{2}$Cornell University\qquad $^{3}$Harvard University \\[4pt]
  \normalfont\small\texttt{\{w564zhan, lhotsko, pynie, yuntian\}@uwaterloo.ca} \\[1pt]
  \normalfont\small\texttt{wk247@cornell.edu}\qquad\texttt{shieber@seas.harvard.edu} \\[4pt]
  \normalfont $^{*}$Equal contribution%
}

\begin{document}

\maketitle

\begin{abstract}
Many everyday programming tasks resist clean rule-based implementation, such as alerting on important log lines, repairing malformed JSON, or ranking search results by intent, and are increasingly outsourced to large language model APIs at the cost of locality, reproducibility, and price. We propose \emph{fuzzy-function programming}: compiling such a function from a natural-language specification into a compact, locally executable neural artifact. We instantiate this paradigm with Program-as-Weights (PAW), in which a 4B compiler trained on FuzzyBench, a 10M-example dataset we release, emits parameter-efficient adapters for a frozen, lightweight interpreter. A 0.6B Qwen3 interpreter executing PAW programs matches the performance of direct prompting of Qwen3-32B, while using roughly one fiftieth of the inference memory and running at 30 tokens/s on a MacBook M3. PAW reframes the foundation model from a per-input \emph{problem solver} into a \emph{tool builder}: invoked once per function definition, it produces a small reusable artifact whose subsequent calls per function application are cheap and offline.
\end{abstract}

\section{Introduction}
\label{sec:intro}

Programming has historically been about writing explicit rules in a formal language designed for the purpose, a programming language. A function is defined by code, and the computer executes it deterministically. For many tasks, this paradigm works beautifully: sorting numbers, processing structured data, computing matrix products. Yet a large class of real-world functions resists precise specification. Consider, for instance, filtering a computer log to alert someone only on the log lines or messages that matter, repairing malformed JSON, or ranking search results by intent. Even apparently ``simple'' tasks, such as writing a regular expression to parse text with many edge cases, prove brittle. Beyond underspecification, real-world inputs are noisy: typos and format drift routinely break hand-written rules and regexes. These are \emph{fuzzy functions}~\citep{10.5555/2594622.2594630}: problems that humans find intuitive but that cannot be fully captured by crisp symbolic rules.

Today, developers frequently outsource such fuzziness to LLM APIs. It is increasingly common to find codebases where a remote LLM is called (e.g., \texttt{gpt(``extract answer'', text)}) to implement functions that are otherwise intractable to program. This approach is undeniably convenient, but it is costly, fragile, undermines reproducibility because providers may silently update their models~\citep{kim-etal-2023-fantom}, and prevents software from being self-contained.

\begin{figure*}[!t]
    \centering
    \includegraphics[width=0.8\linewidth, trim=0 0cm 0 6.3cm]{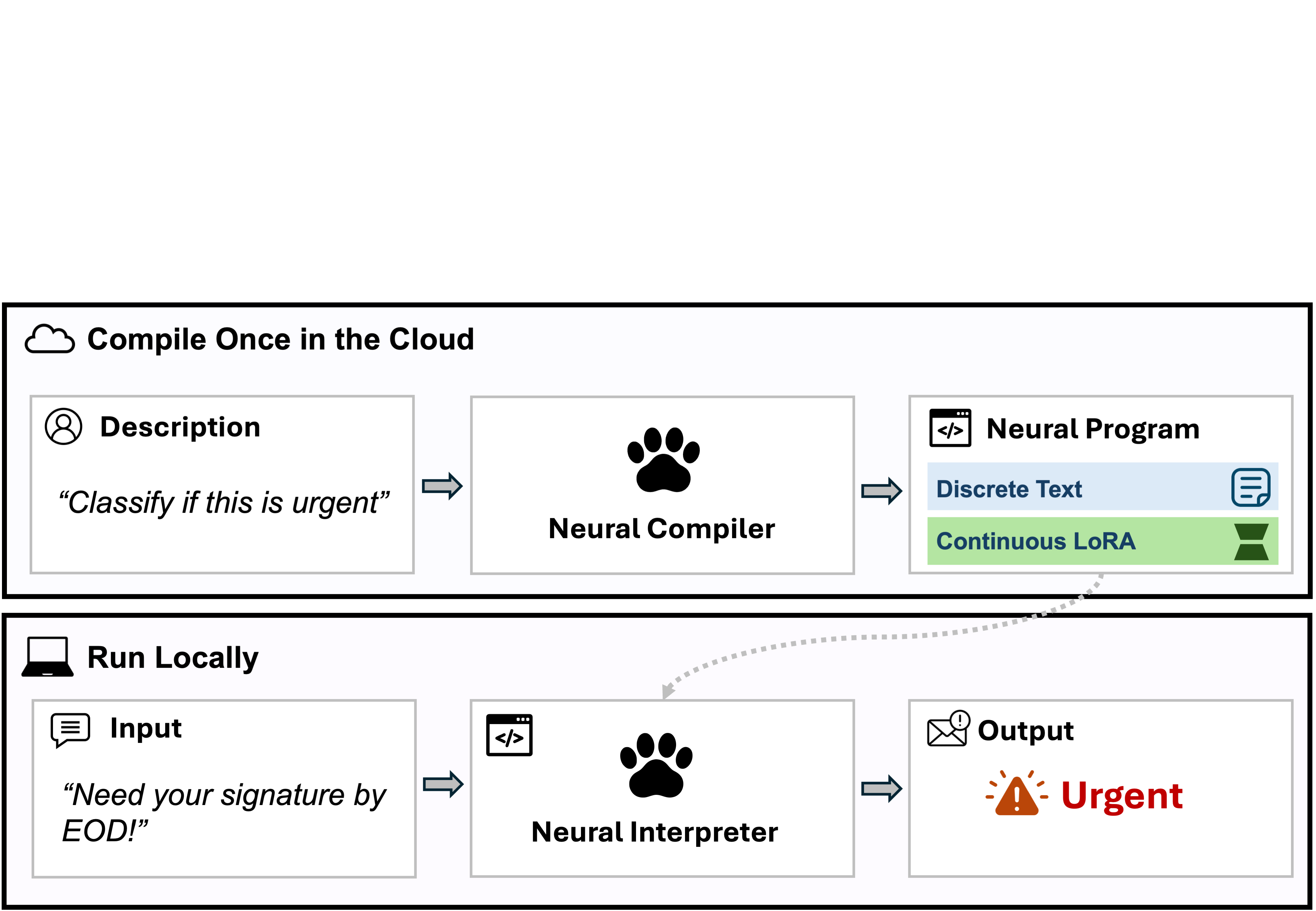}
    \caption{\label{fig:overview}\textbf{Overview of the Program-as-Weights paradigm.} \emph{Top: compile once in the cloud.} A natural-language description of a fuzzy function (here, ``classify if this is urgent'') is fed to a neural compiler, which produces a neural program. \emph{Bottom: run locally.} A small frozen neural interpreter loads the compiled program and runs the user's input (``Need your signature by EOD!'') to produce the output (``urgent''). The compiled program is a single file that can be cached, version-controlled, and called offline like any other library function.}
\end{figure*}

We propose a different paradigm with three steps: the developer \emph{describes} the function in natural language; a neural \emph{compiler} turns that description into a small neural binary; and a frozen, lightweight neural \emph{interpreter}, installed once on the user's device, \emph{runs} that binary just like a user-defined function (\Cref{fig:overview}). We call this paradigm Program-as-Weights (PAW). Any sufficiently expressive parameter-efficient module (PEFT) emitted by a hypernetwork can serve as the program form; we instantiate two, prefix-tuning and text-to-LoRA, and find LoRA better, with future PEFTs possibly better still.

A PAW program has two halves. The first is a \emph{pseudo-program} in natural language, a restatement of the user's specification. The second is a PEFT module that re-tunes the frozen interpreter for this one task: in our precursor system this was a prefix-tuning KV cache; in our current system it is a LoRA generated by the compiler from its own hidden states and injected into the interpreter. The discrete half shields the interpreter from typos and ambiguity in the original specification; the continuous half supplies the fine-grained behavioral control that text alone cannot.

The compile pipeline has two stages, both running 4B Qwen3 models~\citep{yang2025qwen3technicalreport}. The first stage is a \emph{pseudo compiler}, an off-the-shelf model we never train: prompted with a small task-rewriting template, it turns the user's spec into a clean pseudo-program of a paraphrased description plus a handful of input-output examples. The second stage is a \emph{LoRA compiler} that we train: it reads the spec and the pseudo-program and emits the LoRA. We train the LoRA compiler on \textsc{FuzzyBench}, a 10M-example dataset we release with this paper, built incrementally across 29 thematic versions covering more than 800 categories of fuzzy text tasks such as classification, format conversion, parsing, fuzzy matching, natural-language commands, agentic tool use, and many more.

The result is a small, fast, and accurate system. A Qwen3-0.6B interpreter executing PAW programs outperforms direct prompting of Qwen3-32B (73.78\% vs.\ 68.70\% exact match) at roughly one fiftieth the inference memory. Quantized, the same system runs at 30 tokens per second on a MacBook M3 from a $\sim$430~MB GGUF base shared across functions plus a 23~MB per-program LoRA adapter; a smaller GPT-2~\citep{radford2019gpt2} path runs entirely client-side in the browser via WebAssembly.

We see Program-as-Weights as a concrete step toward a small-model future~\citep{belcak2025smalllanguagemodelsfuture}, in which the heavy lifting happens once at compile time and the day-to-day work of running software happens locally.
We illustrate its applications in five case studies: \emph{output triage} (event-driven log monitoring), \emph{custom classification} (intent-based site navigation), \emph{fuzzy search} (semantic search reranking), \emph{agent preprocessing} (a tool-calling pipeline that scores 93\% on \textsc{ToolCall-15}), and \emph{creative generation} (a multilingual word-guessing game). Each is the kind of fuzzy task that resists symbolic implementation but does not need an API call to a 30B-parameter model on every input.
We additionally show the abstraction's modality generality: replacing only the compiler with a vision-language model while keeping the same interpreter runs PAW programs on image-conditioned fuzzy tasks. 
Our code can be found at \url{https://github.com/programasweights} and a public demo is available at \url{https://programasweights.com}.

\section{Programs as Weights}
\label{sec:paradigm}

Let $f : X \to Y$ denote a function whose behavior is more naturally specified through natural language, examples, or constraints than through symbolic code, a \textit{fuzzy function}. Instead of repeatedly invoking an LLM to approximate $f$, we propose to compile a \emph{neural program} that specializes a fixed model to implement $f$.

Formally, let $s$ denote a user specification, expressed in natural language and optionally accompanied by example input-output pairs $(x, y)$. A neural \texttt{Compiler} maps $s$ to a program $p$. A small fixed neural \texttt{Interpreter} executes $p$ on inputs $x \in X$ to produce outputs $\hat{y} \in Y$:
\begin{equation}
\label{eq:paw}
p \;=\; \texttt{Compiler}(s),
\qquad
\hat{y} \;=\; \texttt{Interpreter}(p, x)
\;\approx\; f(x).
\end{equation}
This division mirrors classical programming, where a compiler translates source code into an executable that is later run by a runtime. The crucial difference is that the executable here is a learned parameter blob, and the runtime is a neural network. The interpreter does not need to be retrained: introducing a new fuzzy function only requires compiling a new program $p$.

\paragraph{Hybrid programs.}
For conceptual simplicity, $p$ may be viewed as a single continuous object. In our concrete instantiation, however, $p$ is a hybrid of a discrete and a continuous component:
\begin{equation}
\label{eq:hybrid_program}
p \;=\; \bigl( p_{\text{discrete}}, \; p_{\text{continuous}} \bigr).
\end{equation}
The discrete component $p_{\text{discrete}}$ is a variable-length sequence of tokens that acts as a self-contained ``pseudo-program'' presented to the interpreter as part of its input. The continuous component $p_{\text{continuous}}$ can be implemented using any PEFT method, such as a LoRA injected into the interpreter. %

\paragraph{Why ``program''?}
This framing matters because it determines how the artifact is used. A compiled PAW program is a single file ($\sim$23 MB at $Q4\_0$ for a 0.6B interpreter, plus a one-time shared base) that can be saved, version-controlled, distributed via package managers, and called from Python or JavaScript with a two-line API. PAW programs are objects of the same kind as Python modules: they have a name and a version, but their behavior is encoded in weights rather than in source code. The compiler is the part that does the heavy lifting; the interpreter is a fixed runtime, comparable to a CPU or a byte-code interpreter in conventional software stacks.

\section{The Compiler--Interpreter System}
\label{sec:architecture}

\subsection{Compiler--interpreter abstraction}
\label{sec:arch_abstraction}

The PAW pipeline has three components, none of which depend on the specific PEFT chosen for the program form. A \emph{pseudo compiler} $C_p$ reads the spec $s$ and produces a discrete pseudo-program $p_{\text{discrete}}$. A \emph{PEFT compiler} $C_{\text{PEFT}}$ reads the spec together with $p_{\text{discrete}}$ and emits a small parameter-efficient module $p_{\text{continuous}}$ from its hidden states. The frozen \emph{interpreter} ingests $p_{\text{continuous}}$ at runtime --- by attaching it to the appropriate target modules and running the user's input $x$ through it --- to produce the output $\hat{y}$. We instantiate the PEFT module in two ways: a prefix-tuning KV cache (\Cref{sec:arch_prefix}) and a LoRA (\Cref{sec:arch_lora}), with the latter being our current best.

\paragraph{Pseudo compiler.}
The pseudo compiler $C_p$ is an off-the-shelf Qwen3-4B-Instruct-2507 model that we never train. Given a specification $s$, we prompt $C_p$ with a small task-rewriting template (full text in \Cref{appendix:paw_prompts}) that asks for a clean restatement of the task plus a handful of representative input-output examples. The output is the discrete component $p_{\text{discrete}}$ of the program.\footnote{In an early prototype we trained a single compiler to generate this discrete component via reinforcement learning, but observed that, regardless of seed prompt, the compiler converged on this same paraphrase-plus-examples format; we therefore eliminate the RL stage by directly hand-crafting a prompt that produces this format from an off-the-shelf model.} The pseudo compiler is shared by both PEFT instantiations below.

\subsection{Text-to-LoRA: our current best}
\label{sec:arch_lora}

\begin{figure}[!t]
    \centering
    \includegraphics[width=\linewidth]{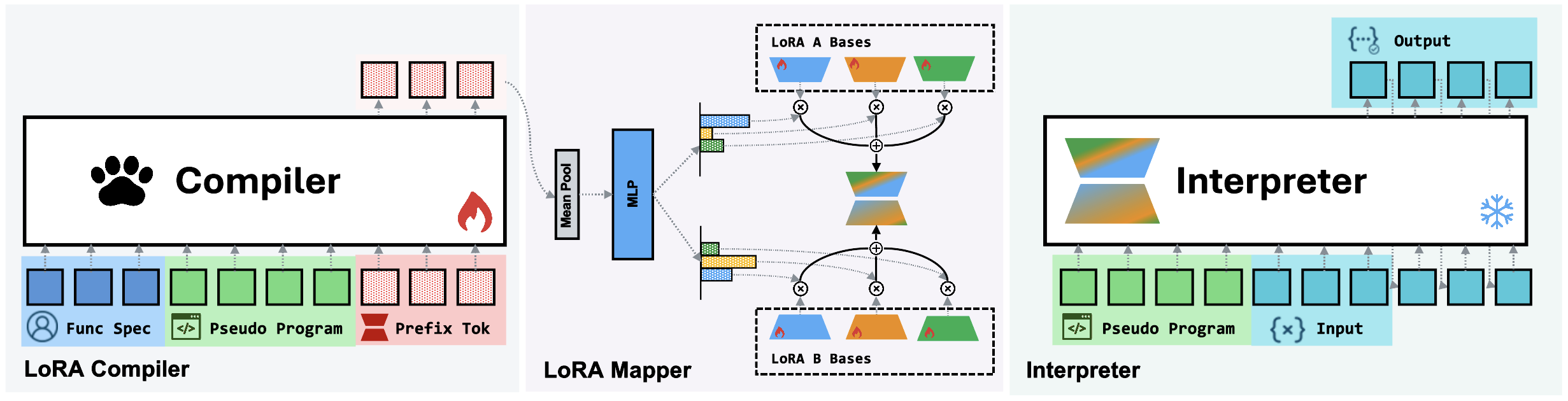}
    \caption{\label{fig:architecture_lora}\textbf{Text-to-LoRA instantiation of PAW (\Cref{sec:arch_lora}).} \emph{Left.} The trained LoRA compiler reads the function specification, the pseudo-program produced by an off-the-shelf prompted pseudo compiler $C_p$ (not depicted), and a fixed sequence of learned prefix tokens; it emits prefix-position hidden states $H$. \emph{Middle.} The LoRA mapper mean-pools $H$, passes it through an MLP, and projects into mixing coefficients that compose LoRA matrices $(A^{\text{ex}}, B^{\text{ex}})$ over shared learnable bases (\cref{eq:lora_mixing}). \emph{Right.} The frozen interpreter ingests $p_{\text{discrete}}$ prepended to the user input $x$, with the LoRA hot-attached, and generates the output autoregressively. The same pipeline holds for the prefix-tuning precursor (\Cref{sec:arch_prefix}, with architecture in \Cref{fig:architecture_prefix}); only the mapping from compiler hidden states to PEFT module changes (LoRA $\to$ KV-cache mapper).}
\end{figure}

\paragraph{LoRA compiler.}
The LoRA compiler $C_L$ is a second 4B Qwen3 model, initialized from the same checkpoint as $C_p$ but \emph{trained} (\Cref{sec:training}). Given the spec $s$ and the pseudo-program $p_{\text{discrete}}$ produced by $C_p$, $C_L$ runs a single forward pass on the concatenation $[s \mid p_{\text{discrete}} \mid \texttt{EOS} \mid \tau_1, \dots, \tau_T]$, where $\tau_{1{:}T}$ is a fixed sequence of $T = 64$ learned ``prefix'' tokens. We extract prefix-position hidden states from $L$ compiler layers spaced uniformly by depth ratio (one per interpreter layer), and stack them into the tensor $H \in \mathbb{R}^{L \times T \times d_{\text{teacher}}}$ that is fed to the LoRA mapper (\Cref{fig:architecture_lora}).

\paragraph{LoRA mapper.}
The LoRA compiler's hidden states $H$ are converted into per-example LoRA weights by a small parameter-efficient module, the \emph{LoRA mapper}. For each interpreter target-module type $m$ (attention $q\!/\!k\!/\!v\!/\!o$ and MLP $\text{gate}/\text{up}/\text{down}$), the mapper maintains shared learnable bases
\[
A^{(m)}_{1{:}N} \;\in\; \mathbb{R}^{N \times r \times d_{\text{in}}^{(m)}},
\qquad
B^{(m)}_{1{:}N} \;\in\; \mathbb{R}^{N \times d_{\text{out}}^{(m)} \times r}.
\]
These hidden states are mean-pooled over both the $L$ depth-aligned layers and the $T$ prefix positions, $\bar{h} = \tfrac{1}{LT}\sum_{l, t} H_{l, t}$, passed through a shallow MLP trunk $\phi$, and projected into mixing coefficients $\alpha^{A,B}_{l,m,n} \in \mathbb{R}$ for each layer $l$, module type $m$, and basis index $n$, via a single linear head. The LoRA at layer $l$ and module $m$ is
\begin{equation}
\label{eq:lora_mixing}
A^{\text{ex}}_{l, m} \;=\; \sum_{n=1}^{N} \alpha^{A}_{l, m, n} \, A^{(m)}_n,
\qquad
B^{\text{ex}}_{l, m} \;=\; \sum_{n=1}^{N} \alpha^{B}_{l, m, n} \, B^{(m)}_n.
\end{equation}
We use rank $r = 64$ and $N = 64$ shared bases per module type, applied to all layers and module types. Per fuzzy function, this injects approximately 38.5M LoRA parameters into the interpreter.\footnote{We compare this design to several more-expressive alternatives (per-position aggregation, per-layer bases, per-position with per-layer bases, LoRA with prefix-tuning) in \Cref{sec:ablations}; none improve over the simple shared-basis design.}

\paragraph{Interpreter.}
The interpreter is a frozen language model. To execute a PAW program on input $x$, we (i) attach the LoRA in \cref{eq:lora_mixing} to the appropriate target modules, (ii) prepend $p_{\text{discrete}}$ to the input $x$, and (iii) generate the output autoregressively. Because the interpreter is frozen and the LoRA hot-swappable, a single device-resident interpreter can serve unboundedly many PAW programs; \Cref{fig:program_library} illustrates this ``one runtime, many programs'' picture for three example specifications.

\subsection{Prefix-tuning: a precursor instantiation}
\label{sec:arch_prefix}

\paragraph{Prefix compiler.}
Our precursor system replaced the LoRA mapper with a \emph{prefix-tuning mapper}. The prefix compiler $C_P$ is a second 4B Qwen3 model trained the same way as $C_L$, with the only difference being how its prefix-position hidden states are consumed. Given $[s \mid p_{\text{discrete}} \mid \texttt{EOS} \mid \tau_{1:T}]$, $C_P$ produces hidden states $H \in \mathbb{R}^{L \times T \times d_{\text{teacher}}}$ at the same $L$ depth-aligned layers as in \Cref{sec:arch_lora}. Instead of pooling and projecting into LoRA weights, a small linear mapper $\psi$ projects these hidden states position-wise into KV pairs $(K^{\text{ex}}_{l, t}, V^{\text{ex}}_{l, t}) \in \mathbb{R}^{2 \times d_{\text{int}}}$ that are prepended to the interpreter's attention KV cache at every layer, in the manner of standard prefix-tuning~\citep{li-liang-2021-prefix} (see \Cref{fig:architecture_prefix} in the appendix for an architecture diagram). The interpreter then runs $x$ through its frozen attention with the additional $T$ prefix-position keys and values visible to every query.

\paragraph{Both methods solve the task.}
At a controlled comparison scale (same amount of training compute), the prefix-tuning instantiation reaches 50.4\% exact match on FuzzyBench, while the LoRA instantiation reaches 56.5\% at $r{=}18$ ($r{=}18$ matches the prefix-tuning program size) and 65.7\% at $r{=}64$ (\Cref{tab:peft_comparison}). Both outperform the no-compiler prompting baseline (9.8\%). LoRA is the stronger PEFT and is the instantiation we scale to the full training data in subsequent experiments. %

\begin{table}[h]
\centering
\caption{\label{tab:peft_comparison}\textbf{Two PEFT instantiations.} Both methods outperform the prompting baseline.}
\begin{tabular}{@{}lc@{}}
\toprule
Method & Accuracy \\
\midrule
Prompting & 0.098 \\
Prefix Tuning        & 0.504 \\
Text-to-LoRA, $r{=}18$                      & 0.565 \\
{Text-to-LoRA, $r{=}64$ (default)}    & \textbf{0.657} \\
\bottomrule
\end{tabular}
\end{table}

\section{Training}
\label{sec:training}

Only the PEFT compiler is trained. The pseudo compiler $C_p$ is held off-the-shelf and frozen; the interpreter is also frozen. The PEFT compiler is trained to produce a PEFT adapter that, when injected into the frozen interpreter alongside a fixed pseudo-program, maximizes the likelihood of the target output. With both endpoints frozen, this reduces to a single supervised objective. We concentrate below on the LoRA instantiation; the prefix-tuning precursor (\Cref{sec:arch_prefix}) was trained with the same SFT recipe described below, substituting the prefix-tuning mapper for the LoRA mapper.

\paragraph{Objective.}
For each training triple $(s, x, y)$, we look up a pre-generated pseudo-program $p_{\text{discrete}} = C_p(s)$, run a forward pass through $C_L$ on $[s \mid p_{\text{discrete}} \mid \texttt{EOS} \mid \tau_{1{:}T}]$ to obtain prefix-position hidden states, pass those through the LoRA mapper to obtain $p_{\text{LoRA}}$, and inject the result into the interpreter. The loss is the negative mean-token log-likelihood of the target $y$ under the frozen interpreter:
\begin{equation}
\label{eq:objective}
\mathcal{L}(\theta) \;=\; \mathbb{E}_{(s,x,y)}\!\left[ - \log P_{\phi}\!\left( y \,\big|\, p_{\text{discrete}},\, p_{\text{LoRA}}(\theta;\, s, p_{\text{discrete}}),\, x \right) \right],
\end{equation}
where $\theta$ denotes the parameters of $C_L$ and the LoRA mapper, and $\phi$ are the interpreter parameters. The gradient flows back through the frozen interpreter into the LoRA mapper and from there into $C_L$'s hidden states. Full hyperparameters and compute setup are in \Cref{appendix:training_details}.

\section{FuzzyBench: A 10M-Example Dataset of Fuzzy Functions}
\label{sec:fuzzybench}

A central obstacle to training PAW-style methods is the lack of a public dataset for ``compile a fuzzy function from a specification.'' We construct \textbf{FuzzyBench}, a 10M-example dataset in which every example is a triple $(s, x, y)$ of (specification, input, target output), generated using \texttt{gpt-5.2}.

\paragraph{Construction.}
We use a two-stage pipeline. In the first stage, we prompt \texttt{gpt-5.2} to generate natural-language specifications of fuzzy functions. Each prompting call produces eight specifications, and we run repeated calls under different category constraints to cover the breadth of fuzzy tasks developers actually encounter. In the second stage, for each specification, we prompt \texttt{gpt-5.2} again to generate eight input/output pairs. Specifications are split 80/10/10 by spec into train/validation/test, so that test specifications are entirely unseen at training time. For evaluation, we construct a \emph{verified} test set on which an independent strong model (\texttt{gpt-5-mini}; \citealp{openai2025gpt5}) and \texttt{gpt-5.2} agree on the output, removing examples where the target itself is ambiguous. Full prompts are in \Cref{appendix:prompts}.

\begin{figure}[!t]
\centering
\includegraphics[width=0.85\linewidth]{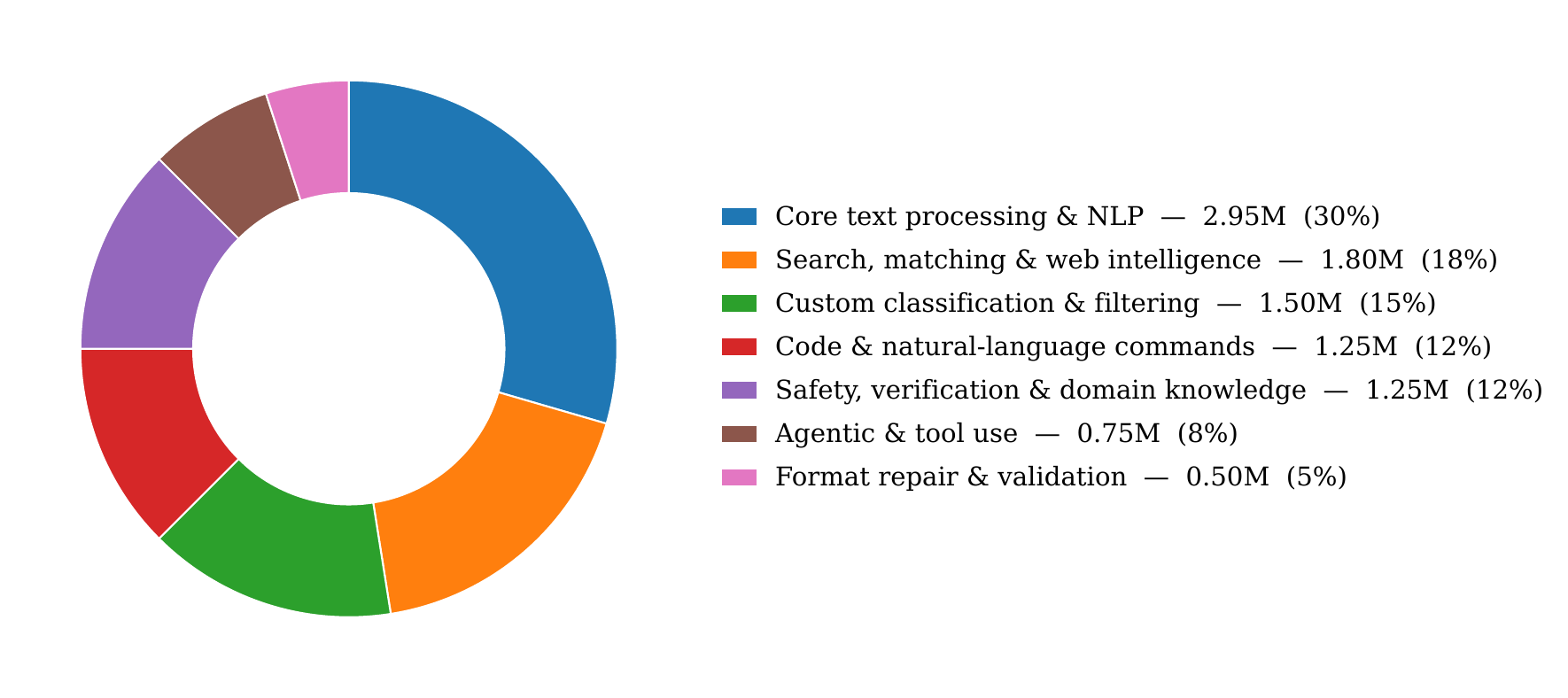}
\caption{\label{fig:fuzzybench_categories}\textbf{FuzzyBench-10M task-family distribution.} 29 incremental thematic versions are mapped to 7 high-level families. ``Core text processing \& NLP'' is the largest family because the v1 base layer (2.5M examples; 277 base categories) covers parsing, classification, NER, coreference, and sentiment; the remaining 7.5M examples spread across the other six families.}
\end{figure}

\paragraph{Thematic coverage.}
FuzzyBench is built incrementally over 29 versions, each adding 100K-500K examples covering a new family of fuzzy tasks. \Cref{fig:fuzzybench_categories} groups the resulting 10M examples into seven high-level task families that span what developers actually encounter in fuzzy logic: from raw text processing and parsing to agentic tool use, web intelligence, code-and-command generation, and safety/verification. The full per-version timeline (29 entries; the first version alone establishes 277 base categories, and the final dataset covers more than 800 sub-categories) is in \Cref{appendix:dataset_versions}.

\paragraph{Noise variants.}
For robustness evaluation, we additionally release noise-perturbed versions of the test set along eight axes: typos, grammar errors, ambiguity, formatting drift, ``all noise'' (combined), terse phrasing, casual phrasing, and paraphrase. Each axis comes in three intensity levels (light, medium, heavy). \Cref{sec:noise} reports robustness numbers.

\paragraph{Human validation.}
On 100 random test examples, human answers from original specifications matched 92 reference outputs; 98 descriptions were judged unambiguous (\Cref{appendix:human_validation}).

\section{Main Results}
\label{sec:experiments}

PAW improves over prompting the same 0.6B base on all four external datasets, including accuracy gains from 52.80\% to 90.40\% on YouTube and from 66.88\% to 90.64\% on IMDB (\Cref{tab:main_text}).

\paragraph{Baselines.}
\emph{(i) Direct prompting} of open-weight LMs (Qwen3 0.6B, 4B, 8B, 14B, 32B; OLMo3-7B~\citep{olmo2025olmo3}; gpt-oss-20B~\citep{openai2025gptoss}), and of two API models that bound the empirical ceiling (\texttt{gpt-5-mini} and \texttt{gpt-5.2}). \emph{(ii) Symbolic code generation}: ALCHEmist's LM-to-code pipeline~\citep{NEURIPS2024_72802bef}, which generates Python functions that execute the fuzzy task at inference (\Cref{appendix:python_failure}). \emph{(iii) Standard adaptation of the same 0.6B base}: full finetuning across 1-4 epochs, and fixed (non-compiler-generated) LoRAs at ranks $r \in \{18, 64, 128\}$.

\paragraph{Main result.}
PAW's 0.6B interpreter outperforms Qwen3-32B prompting on FuzzyBench (\textbf{73.78\%} vs.\ 68.70\% EM; \Cref{tab:main_text}) at roughly $50\times$ less inference memory. On 1,618 user-submitted specifications, it achieves 50.68\% versus 51.85\% for Qwen3-32B (\Cref{appendix:user_specs}). %
\begin{table*}[!t]
\centering
\caption{\label{tab:main_text}\textbf{Main results.} FuzzyBench reports verified-test EM; unfiltered results appear in Appendix~\ref{appendix:unfiltered_test}. Following the WRENCH benchmark setup of ALCHEmist~\citep{NEURIPS2024_72802bef}, SMS uses F1 and the rest use Acc. \emph{Contained} indicates the program is self-contained and executable without internet access. PS is per-program shipping size; for prompting baselines this is the prompt/spec size, and for PAW it is the deployed PEFT adapter ($Q4\_0$ quantized for Qwen3 0.6B and Qwen3.5 0.8B; fp32 for GPT-2). $^\dagger$: numbers taken from \citet{NEURIPS2024_72802bef}, which uses 10-sample majority voting; the reimplementation row uses single-sample for fairness. $^*$: zero F1 due to zero recall.}
\renewcommand{\arraystretch}{1.1}
\setlength{\tabcolsep}{3.5pt}
\small
\resizebox{\textwidth}{!}{
\begin{tabular}{@{}l@{}c@{ }c cc cc cc cc cc@{}}
\toprule
& \multirow{2}{*}{Contained} & \multirow{2}{*}{Interp.\ Size} &
\multicolumn{2}{c}{FuzzyBench} &
\multicolumn{2}{c}{YouTube} &
\multicolumn{2}{c}{SMS} &
\multicolumn{2}{c}{Yelp} &
\multicolumn{2}{c}{IMDB} \\
\cmidrule(lr){4-5}\cmidrule(lr){6-7}\cmidrule(lr){8-9}\cmidrule(lr){10-11}\cmidrule(lr){12-13}
Method & & & Acc & PS & Acc & PS & F1 & PS & Acc & PS & Acc & PS \\
\midrule
gpt-5.2 (API)                              & $\times$ & --   & 96.09\% & 0.73 KB & 95.20\%     & 0.95 KB & 97.06\%     & 1.03 KB & 98.55\%     & 1.69 KB & 95.60\%     & 2.25 KB \\
gpt-5-mini (API)                           & $\times$ & --   & 91.87\% & 0.73 KB & 93.60\% & 0.95 KB & 91.03\% & 1.03 KB & 98.13\% & 1.69 KB & 94.96\% & 2.25 KB \\
Local LM (Qwen3 0.6B)                     & $\checkmark$ & 0.6B & 9.84\%  & 0.73 KB & 52.80\% & 0.95 KB & 0.00\%$^*$ & 1.03 KB & 89.55\% & 1.69 KB & 66.88\% & 2.25 KB \\
Local LM (Qwen3 4B)                       & $\checkmark$ & 4B   & 49.63\% & 0.73 KB & 90.80\% & 0.95 KB & 92.54\% & 1.03 KB & 97.53\% & 1.69 KB & 93.76\% & 2.25 KB \\
Local LM (Qwen3 8B)                       & $\checkmark$ & 8B   & 52.15\% & 0.73 KB & 94.40\% & 0.95 KB & 91.55\% & 1.03 KB & 97.95\% & 1.69 KB & 94.52\% & 2.25 KB \\
Local LM (Qwen3 14B)                      & $\checkmark$ & 14B  & 63.96\% & 0.73 KB & 93.20\%     & 0.95 KB & 92.75\%     & 1.03 KB & 97.74\%     & 1.69 KB & 92.64\%     & 2.25 KB \\
Local LM (Qwen3 32B)                      & $\checkmark$ & 32B  & 68.70\% & 0.73 KB & 93.60\%     & 0.95 KB & 89.04\%     & 1.03 KB & 98.11\%     & 1.69 KB & 94.64\%     & 2.25 KB \\
Local LM (OLMo3 7B)                       & $\checkmark$ & 7B   & 39.84\%     & 0.73 KB & 90.00\% & 0.95 KB & 90.14\% & 1.03 KB & 97.66\% & 1.69 KB & 93.28\% & 2.25 KB \\
Local LM (gpt-oss-20B)        & $\checkmark$ & 20B  & 85.45\% & 0.73 KB & 91.60\%     & 0.95 KB & 89.05\%     & 1.03 KB & 97.42\%     & 1.69 KB & 92.08\%     & 2.25 KB \\
LM$\to$Code~\citep{NEURIPS2024_72802bef}  & $\checkmark$ & 29 MB & --     & --      & 89.10\%$^\dagger$ & -- & 90.00\%$^\dagger$ & -- & 57.50\%$^\dagger$ & -- & 66.20\%$^\dagger$ & -- \\
LM$\to$Code (Reimplementation)            & $\checkmark$ & 29 MB & 35.81\% & 0.08 KB & 70.46\% & 0.09 KB & 86.41\% & 0.06 KB & 50.35\% & 0.05 KB & 73.92\% & 0.08 KB \\
\midrule
\textbf{PAW (Qwen3 0.6B)}                  & $\checkmark$ & 0.6B  & {73.78\%} & 23 MB & 90.40\% & 23 MB & 80.77\% & 23 MB & 95.82\% & 23 MB & 90.64\% & 23 MB \\
\textbf{PAW (Qwen3.5 0.8B)}                & $\checkmark$ & 0.8B  & 67.29\% & 16 MB & 88.40\% & 16 MB & 84.55\% & 16 MB & 94.05\% & 16 MB & 82.68\% & 16 MB \\
\textbf{PAW (GPT-2 124M)}                  & $\checkmark$ & 124M  & 54.39\% & 38 MB & 93.60\% & 38 MB & 77.50\% & 38 MB & 93.16\% & 38 MB & 82.12\% & 38 MB \\
\bottomrule
\end{tabular}}
\end{table*}

\paragraph{Beyond generated Python.}
On 100 sampled FuzzyBench specifications with four test examples each, generated Python failed all four examples for 33 specifications. PAW achieved 63.6\% exact match on those same 132 examples, demonstrating the value of neural programs for inputs missed by generated symbolic code (Appendix~\ref{appendix:python_failure}).

\paragraph{Cross-interpreter scaling.}
Among three interpreters GPT-2 124M, Qwen3 0.6B, and Qwen3.5 0.8B~\citep{qwen2026qwen35}, Qwen3 0.6B is the strongest interpreter; the hybrid 0.8B is slightly weaker. GPT-2 124M, despite having only 1/5 the parameters of Qwen3 0.6B and no instruction tuning, still achieves 54\%, suggesting that the compiler-generated LoRA can encode usable task adaptations even into very small, weakly-capable bases. %

\paragraph{Multimodal generalization without changing the interpreter.}
The compiler-interpreter abstraction extends to image-conditioned fuzzy functions \emph{without} changing the interpreter. We swap the text-only Qwen3-4B-Instruct compiler for the same-family Qwen3-VL-4B compiler~\citep{Qwen3-VL2025}, keep the same Qwen3 0.6B interpreter, and reuse the same LoRA mapper. Image information is conveyed through the compiled pseudo-program and PEFT module, so the small text interpreter never sees pixels. \Cref{tab:main_image} reports six image-conditioned tasks: three CoSyn-400K diagram-understanding tasks (Chemical, Circuit, Music)~\citep{yang2025scaling,deitke2024molmo}, the structured-output Im2LaTeX-100K~\citep{pmlr-v70-deng17a} and Im2SMILES-20K~\citep{deng2023markuptoimage} tasks, and the open-ended visual question answering TextVQA~\citep{singh2019textvqa}; full prompts are in \Cref{appendix:paw_image_prompts}.

PAW (LoRA) outperforms all VLM baselines (up to 4B parameters) on the three CoSyn diagram tasks (Circuit 0.274 vs.\ 0.196 best baseline; Chemical 0.414 vs.\ 0.258; Music 0.552 vs.\ 0.470) at $\sim$0.6B interpreter size. On the long-form structured generation task Im2LaTeX, PAW (LoRA) is weaker than its prefix-tuning precursor  (0.181 vs.\ 0.391); a discrete-pseudo-only ablation in \Cref{appendix:image_ablations} shows the gap arises because the long input/output examples in the pseudo-program crowd the small interpreter's context budget on long-form tasks.

\begin{table*}[!t]
\centering
\caption{\label{tab:main_image}\textbf{Image-conditioned fuzzy functions.} The PAW rows use the same Qwen3 0.6B and Qwen 3.5 0.8B interpreters as in \Cref{tab:main_text} (only the compiler is swapped, from Qwen3-4B-Instruct to Qwen3-VL-4B).}
\renewcommand{\arraystretch}{1.1}
\setlength{\tabcolsep}{4pt}
\small
\resizebox{\textwidth}{!}{
\begin{tabular}{@{}lc cccccc@{}}
\toprule
Method & Interp.\ Size & Circuit & Chemical & Music & Im2SMILES & Im2LaTeX & TextVQA \\
\midrule
AndesVL 0.6B~\citep{jin2025andesvltechnicalreportefficient} & 0.6B  & 0.183 & 0.214 & 0.448 & 0.000 & 0.435 & 0.718 \\
Qwen3-VL 2B-Instruct~\citep{Qwen3-VL2025}                   & 2B    & 0.186 & 0.258 & 0.470 & 0.016 & 0.408 & {0.836} \\
Qwen3-VL 4B-Instruct~\citep{Qwen3-VL2025}                   & 4B    & 0.196 & 0.221 & 0.450 & 0.044 & 0.399 & 0.822 \\
\midrule
\textbf{PAW prefix-tuning (Qwen3 0.6B)}     & 0.6B  & 0.241 & 0.365 & 0.525 & 0.175 & {{0.391}} & 0.612 \\
\textbf{PAW LoRA (Qwen3 0.6B)}                         & 0.6B  & {{0.274}} & {{0.414}} & {{0.552}} & 0.203 &  0.181 & {0.721} \\
\textbf{PAW LoRA (Qwen3.5 0.8B)}                         & 0.8B  & {{0.284}} & {{0.438}} & {{0.573}} & {{0.285}} & 0.204 & {0.755} \\
\bottomrule
\end{tabular}}
\end{table*}

\section{Ablations}
\label{sec:ablations}

\paragraph{Architectural variants of the LoRA mapper.}
\label{sec:ablation_mapper_design}
We tried several variants of the LoRA mapper that, on paper, are strictly more expressive than the default (mean-pool over prefix tokens, shallow trunk, shared bases). Each made things worse. \Cref{tab:ablation_mapper} reports accuracy across these variants. The simplest design of mean-pooling the prefix-token hidden states into one vector, running a single residual MLP, and projecting to mixing coefficients over a shared basis set, is the strongest. We do not have a clean theoretical explanation for this; we report the finding so that future work need not rediscover it.

\begin{table}[!t]
\centering
\begin{minipage}[t]{0.59\linewidth}
\centering
\caption{\label{tab:ablation_mapper}\textbf{Architectural variants of the LoRA mapper.} ``More expressive'' design choices that we expected to help all underperformed the simple default.}
\small
\setlength{\tabcolsep}{4pt}
\begin{tabular}{@{}lc@{}}
\toprule
Mapper variant & Accuracy \\
\midrule
\textbf{Default} ($r{=}64$, $N{=}64$, shared bases) & \textbf{0.6223} \\
Per-position aggregation                          & 0.5598 \\
Per-position $+$ per-layer bases                  & 0.5559 \\
Per-layer bases (only)                            & 0.6028 \\
LoRA $+$ prefix-tuning (both pathways)            & 0.6033 \\
\bottomrule
\end{tabular}
\end{minipage}\hfill
\begin{minipage}[t]{0.39\linewidth}
\centering
\caption{\label{tab:killer_baselines}\textbf{No compiler baselines.} PAW row reproduced from \Cref{tab:main_text} for comparison.}
\small
\setlength{\tabcolsep}{6pt}
\begin{tabular}{@{}lc@{}}
\toprule
Method (0.6B base) & Accuracy \\
\midrule
Fixed LoRA $r{=}18$         & 0.4236  \\
Fixed LoRA $r{=}64$          & 0.5210 \\
Fixed LoRA $r{=}128$         & 0.5159 \\
Full finetuning          & 0.5840  \\
\textbf{PAW (Qwen3 0.6B)}             & \textbf{0.7378}  \\
\bottomrule
\end{tabular}
\end{minipage}
\end{table}

\paragraph{Compiler vs.\ no compiler.}
\Cref{tab:killer_baselines} compares PAW with full finetuning and fixed LoRAs at three ranks on the same Qwen3 0.6B base, data, and training budget. PAW gains 15.4 percentage points over full finetuning and 21.7 over the strongest fixed LoRA. Its advantage over full finetuning holds across all six represented task families, with gains of 11.6--21.4 points (\Cref{appendix:task_families}), demonstrating broad benefits from compiler-generated LoRA.

\paragraph{Other ablations.}
\label{sec:ablations_misc}
Additional ablations on model architecture decisions can be found in \Cref{appendix:ablations}.

\section{Robustness to Noisy Specifications}
\label{sec:noise}

Real specifications written by developers are noisy: they contain typos, ambiguity, and grammar errors. We evaluate PAW on noise-perturbed versions of the test\_clean specifications across seven axes (typos, grammar, ambiguity, formatting, all-noise combined, terse, paraphrase).

\Cref{tab:noise} reports the robustness results. PAW degrades only slightly under heavy noise --- but \emph{why}? We hypothesise that this robustness is mediated by the discrete pseudo-program: the 4B compiler converts the noisy spec into a clean restatement before the small interpreter ever sees it. To test the hypothesis, we trained a variant that bypasses the pseudo-program and feeds the raw spec $s$ directly to the interpreter. The result (\Cref{tab:ablation_pseudo_denoiser}) confirms the hypothesis. On clean inputs, feeding the pseudo-program rather than the raw spec is only 1.6 points better. On \emph{heavy-typo} specifications, however, the gap widens to 4.5 points. The compiler, a 4B LM whose entire job is to read fuzzy specifications and emit a clean restatement, effectively denoises the input that the small interpreter sees, which is why PAW degrades little when the original specification is corrupted.

\begin{table}[!t]
\centering
\caption{\label{tab:noise}\textbf{Robustness to noise.} The seven noise variants shown modify the spec but leave the input unchanged. PAW degrades only slightly even under combined heavy noise.}
\small
\setlength{\tabcolsep}{4pt}
\begin{tabular}{@{}lcccccccc@{}}
\toprule
& clean & typos & grammar & ambiguity & format & all-noise & terse & paraphrase \\
\midrule
PAW Qwen3 0.6B (e2) & 0.6692 & 0.6621 & 0.6731 & 0.6511 & 0.6526 & 0.6326 & 0.6499 & 0.6614 \\
\quad drop from clean & --     & $-0.7\%$ & $+0.4\%$ & $-1.8\%$ & $-1.7\%$ & $-3.7\%$ & $-1.9\%$ & $-0.8\%$ \\
\bottomrule
\end{tabular}
\end{table}

\begin{table}[!t]
\centering
\caption{\label{tab:ablation_pseudo_denoiser}\textbf{The pseudo-program protects the interpreter from noisy specifications.} On heavy-typo specifications, feeding raw spec is 4.5 points worse than feeding the pseudo-program to the interpreter.}
\small
\setlength{\tabcolsep}{8pt}
\begin{tabular}{@{}lcc@{}}
\toprule
Interpreter input & Accuracy (clean) & Accuracy (heavy typos) \\
\midrule
Pseudo-program (default)  & 0.6443 & 0.6108 \\
Raw spec                  & 0.6285 & 0.5662 \\
\bottomrule
\end{tabular}
\end{table}

\section{Local Execution}
\label{sec:execution}

PAW compiles functions in 2.33 seconds on our experiment server (\Cref{appendix:compilation_latency}).

\paragraph{Developer interface.}
A PAW program is a single file that can be downloaded, cached, and called via a small Python or JavaScript API. \Cref{fig:paw-api} shows a complete minimal Python example: \texttt{paw.compile(prompt)} sends a specification to a compiler service and returns a serializable program object; \texttt{paw.function(id\_or\_path)} loads a compiled program and exposes it as a Python callable. After the first download, all execution happens locally with no external API calls.

\begin{figure*}[!t]
  \centering
  \begin{minipage}[t]{0.48\textwidth}
  \vspace{0pt}
  \begin{lstlisting}[language=Python,
      basicstyle=\ttfamily\small,
      keywordstyle=\color{blue!60!black},
      commentstyle=\color{green!50!black},
      stringstyle=\color{orange!70!black},
      showstringspaces=false,
      frame=single,
      caption={Compile a fuzzy function},
      captionpos=b]
import programasweights as paw

# (1) Describe the fuzzy function
spec = """Classify if this email  
requires immediate attention.
""".strip()

# (2) Compile to a neural program
program = paw.compile(spec, \
  slug="email-triage")
  \end{lstlisting}
  \end{minipage}
  \hfill
  \begin{minipage}[t]{0.48\textwidth}
  \vspace{0pt}
  \begin{lstlisting}[language=Python,
      basicstyle=\ttfamily\small,
      keywordstyle=\color{blue!60!black},
      commentstyle=\color{green!50!black},
      stringstyle=\color{orange!70!black},
      showstringspaces=false,
      frame=single,
      caption={Run the compiled program locally},
      captionpos=b]
import programasweights as paw

# (3) Load the compiled program
fn = paw.function("email-triage")

# (4) Call it like a function
print(fn("Thesis defense moved "
         "to 3pm; need your "
         "signature today."))
# -> "immediate"
  \end{lstlisting}
  \end{minipage}
  \caption{\label{fig:paw-api}\textbf{Developer interface.} \emph{Left}: the compiler translates a natural-language specification into a neural program. \emph{Right}: the interpreter loads this program and exposes it as a local function.}
\end{figure*}

\paragraph{Quantization without measurable accuracy loss.}
On-device execution requires a small footprint. We quantize both the shared interpreter base and each per-program LoRA adapter to GGUF formats compatible with \texttt{llama.cpp}~\citep{llamacpp}. \Cref{tab:quant_main} reports our quantization findings on the 0.6B Qwen3 interpreter, evaluated on a 4,096-example test subset: a 4-bit base ($Q4\_K\_M$, $\sim$484~MB) plus a $Q4\_0$ LoRA adapter ($\sim$23~MB per program) loses only 1.3 points relative to bf16, and a $Q6\_K$ base plus $Q4\_0$ adapter is statistically indistinguishable from bf16.

\begin{table}[!t]
\centering
\caption{\label{tab:quant_main}\textbf{Quantization on the 0.6B Qwen3 interpreter.} A $Q6\_K$ base + $Q4\_0$ LoRA is indistinguishable from bf16 within noise; $Q4\_K\_M$ loses 1.3 points but cuts total disk to $\sim$507~MB.}
\small
\setlength{\tabcolsep}{6pt}
\begin{tabular}{@{}lccc@{}}
\toprule
Configuration & Base size & Adapter size & Accuracy \\
\midrule
PyTorch bf16 (no quantization)        & 1515 MB & --   & 0.6580 \\
fp16 base + fp32 LoRA                 & 1509 MB & 162 MB & 0.6594 \\
$Q8\_0$ base + $Q4\_0$ LoRA           &  805 MB &  23 MB & 0.6567 \\
$Q6\_K$ base + $Q4\_0$ LoRA           &  623 MB &  23 MB & 0.6575 \\
$Q5\_K\_M$ base + $Q4\_0$ LoRA        &  551 MB &  23 MB & 0.6477 \\
$Q4\_K\_M$ base + $Q4\_0$ LoRA        &  484 MB &  23 MB & 0.6453 \\
$\textsc{IQ4\_XS}$ base + $Q4\_0$ LoRA &  430 MB &  23 MB & 0.6462 \\
\bottomrule
\end{tabular}
\end{table}

\paragraph{Latency on a MacBook M3.}
On a MacBook M3 with Metal acceleration, the $Q5\_K\_M$ base + $Q4\_0$ adapter runs at 31.6 tokens/s with a 0.48 s cold load. Full per-quant tables for GPT-2 124M and Qwen3.5 0.8B are in \Cref{appendix:quant_full}.

\paragraph{Case studies.}
\label{sec:case_studies}
We applied PAW to five use cases. \emph{Event-driven log monitoring} replaces the naive \texttt{wait}-based terminal watching in Cursor with a local classifier that fires only on the lines that matter. \emph{Intent-based site navigation} provides a natural-language quick-find for a website without an LLM API call per request. \emph{Semantic search reranking} adds intent-aware fuzzy search to an existing keyword index, again without putting an LLM in the request path. For \emph{tool calling}, a 10-PAW-function pipeline scores 93\% on \textsc{ToolCall-15}, capturing tool-routing behavior usually reserved for much larger models. The \emph{multilingual word-guessing game} (Alien-Taboo) is a fuzzy interactive game in which each player turn is served by a 0.6B PAW interpreter on a small server, with one PAW program per language; the LLM is invoked only at compile time, which is what makes a game of this kind economical to host. Full details are in \Cref{appendix:case_studies}.

\section{Related Work}
\label{sec:related}

\paragraph{Hypernetworks.}
Hypernetworks~\citep{ha2017hypernetworks} generate a target network's weights. Text-conditioned variants include HINT~\citep{ivison-etal-2023-hint}, which converts instructions and examples into parameter-efficient modules, and HyperTuning~\citep{phang2023hypertuning}, which generates prefixes or LoRAs from few-shot examples. Generative Adapter~\citep{chen2025generative} uses a self-supervised adapter generator to map context into low-rank adaptations for a frozen language model in a single forward pass. Text-to-LoRA~\citep{charakorn2025texttolora} learns to reconstruct pretrained task adapters from textual descriptions; SHINE~\citep{shine2026} and Doc-to-LoRA~\citep{doctolora2026} map contexts into LoRAs for subsequent querying. PAW trains its compiler directly on specification--input--output triples through a frozen interpreter. It compiles a function definition into a hybrid program: a discrete pseudo-program supplies the task description and examples, while continuous PEFT weights specialize the interpreter. Trained on FuzzyBench-10M's programmer-style specifications, the compiler produces reusable functions exposed through a developer API for local execution, versioning, and distribution. Further comparisons appear in Appendix~\ref{appendix:hypernetworks}.

\paragraph{Parameter-efficient finetuning.}
Parameter-efficient finetuning learns compact adaptations to frozen models through adapters~\citep{houlsby2019adapters}, prefixes~\citep{li-liang-2021-prefix}, or low-rank weight updates~\citep{hu2022lora}. PAW's trained compiler generates the PEFT module from a function specification in a forward pass, making new functions available without per-function optimization. Further background appears in Appendix~\ref{appendix:peft_background}.

\paragraph{Synthetic instruction-data generation.}
FuzzyBench-10M builds on LLM-generated instruction datasets such as Self-Instruct~\citep{wang2023selfinstruct}, Unnatural Instructions~\citep{honovich2023unnatural}, and Magpie~\citep{xu2025magpie}. It targets programmer-style fuzzy functions through 29 thematic versions and constructs a verified test split using agreement between two strong models. Appendix~\ref{appendix:synthetic_data_background} provides the detailed comparison.

\paragraph{Model distillation.}
ALCHEmist~\citep{NEURIPS2024_72802bef} distills labelling logic from LLMs into Python programs that run on a standard interpreter. PAW shares the motivation of amortizing LLM usage but compiles directly into \emph{neural} weights instead of textual code, which lets it implement fuzzy functions that resist symbolic encoding. Binder~\citep{cheng2023binder} translates a task input into a SQL/Python program with embedded LM API calls; PAW differs in that the program is the weights themselves, not a piece of text containing API calls. Distilling Step-by-Step~\citep{hsieh2023distillingstepbystep} distills LLM reasoning into smaller finetuned models with rationales as auxiliary supervision; PAW shares the goal of replacing large-LM inference with small-model inference, but achieves it via per-task compile rather than per-task finetuning. Follow-up work, \emph{Compile by Training}~\citep{deng-etal-2026-compile}, initializes each adapter with PAW's fast compiler and then refines it through per-function finetuning on teacher-generated examples. It retains PAW's program format and shared frozen interpreter, extending the same programming abstraction to additional optimization at compile time.

\paragraph{Neural programs.}
Neural networks have long represented and executed programs~\citep{graves2014neuralturingmachines,reed2016neuralprogrammerinterpreters}, including compilation of formal code into weights~\citep{weiss2021thinkingliketransformers,GRUAU19951}. PAW produces reusable programs on demand from natural-language specifications and executes them on a fixed local interpreter. A compiler configures the small model's behavior for each function. Appendix~\ref{appendix:neural_programs_background} provides the detailed comparison.

\section{Conclusion}
\label{sec:conclusion}

We introduced Program-as-Weights, a programming paradigm in which a fuzzy function is compiled once into a small neural binary and executed locally on a fixed interpreter. On FuzzyBench, a 0.6B-parameter interpreter executing PAW programs matches Qwen3-32B prompting at $\sim$50$\times$ less inference memory and runs at 30~tok/s on a MacBook M3 with quantized GGUF; we illustrate the paradigm through five case studies. The same abstraction extends to image-conditioned fuzzy tasks by swapping only the compiler for a vision-language model. We hope Program-as-Weights contributes to a future in which small LMs serve as the runtime~\citep{belcak2025smalllanguagemodelsfuture}, where large models compile and small models execute, and the role of foundation models shifts from per-input \emph{problem solver} to per-function \emph{tool builder}.

\clearpage
\subsubsection*{Acknowledgments}
We thank Sasha Rush for his guidance and contributions to the earlier project that laid the foundation for this work. We also thank Saarang Agarwal, Austin Dong, Mohammad Jaffer Iqbal, Bihui Jin, Yinxi Li, Jiale Amber Wang, and the anonymous reviewers for their valuable comments and feedback.

We thank Braden Hancock at the Laude Institute for his support and for arranging access to computing resources provided by Lambda. This research was generously supported by a Starter Grant from the University of Waterloo, and by the Natural Sciences and Engineering Research Council of Canada (NSERC) under grant numbers RGPIN-2024-04909 and RGPIN-2024-05178. Computational resources were also provided by Compute Ontario (computeontario.ca) and the Digital Research Alliance of Canada (alliancecan.ca). We also thank OpenAI's Research Access Program for providing API credits. Wentao Zhang was supported in part by these sources and by the Dr. Derick Wood Graduate Scholarship, generously funded by Ms. Mary Chen.

\bibliographystyle{plainnat}
\bibliography{references}

\newpage
\appendix

\section{Web Interface for PAW Compilation}
\label{appendix:web_ui}

\begin{figure*}[!ht]
  \centering
  \includegraphics[width=0.9\linewidth]{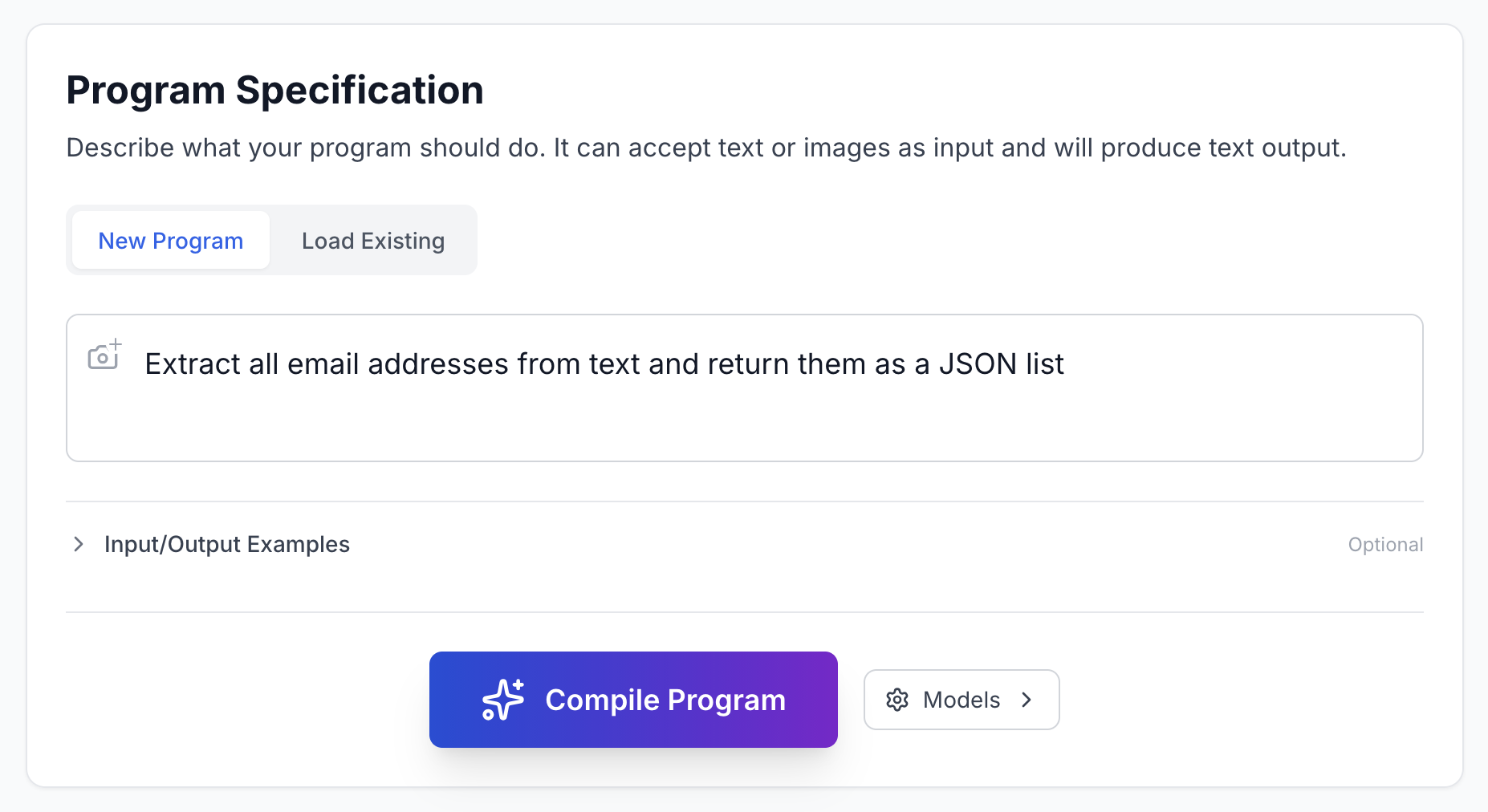}
  \caption{\label{fig:web-ui-1}\textbf{Step 1: Compile a program from natural language.} The user specifies a fuzzy function in natural language. Image inputs are also supported.}
\end{figure*}

\begin{figure*}[!ht]
  \centering
  \includegraphics[width=0.9\linewidth]{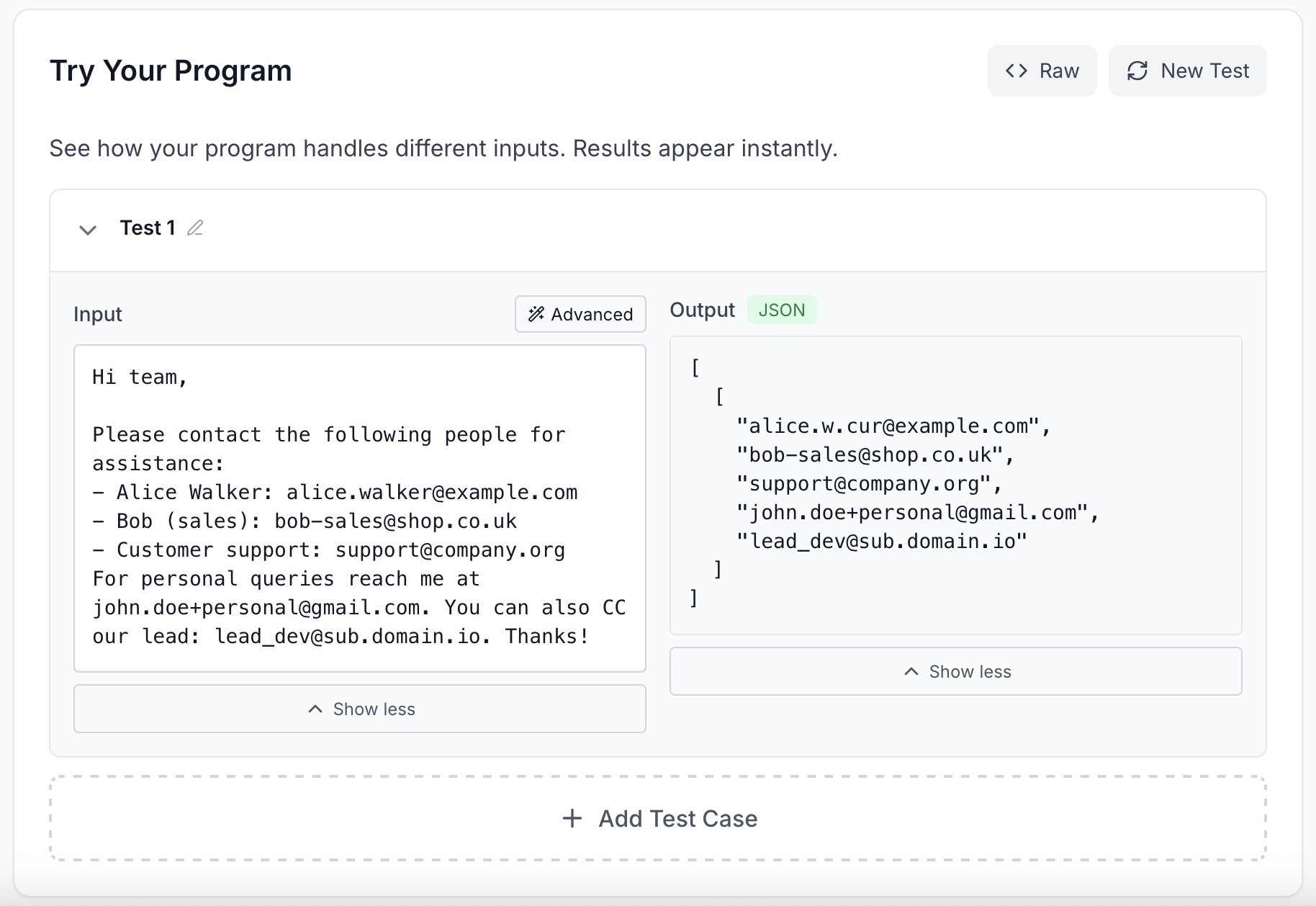}
  \caption{\label{fig:web-ui-2}\textbf{Step 2: Interactively test the compiled program.} Users can provide test inputs and inspect the corresponding outputs, enabling rapid validation and refinement before download.}
\end{figure*}

\begin{figure*}[!ht]
  \centering
  \includegraphics[width=0.9\linewidth]{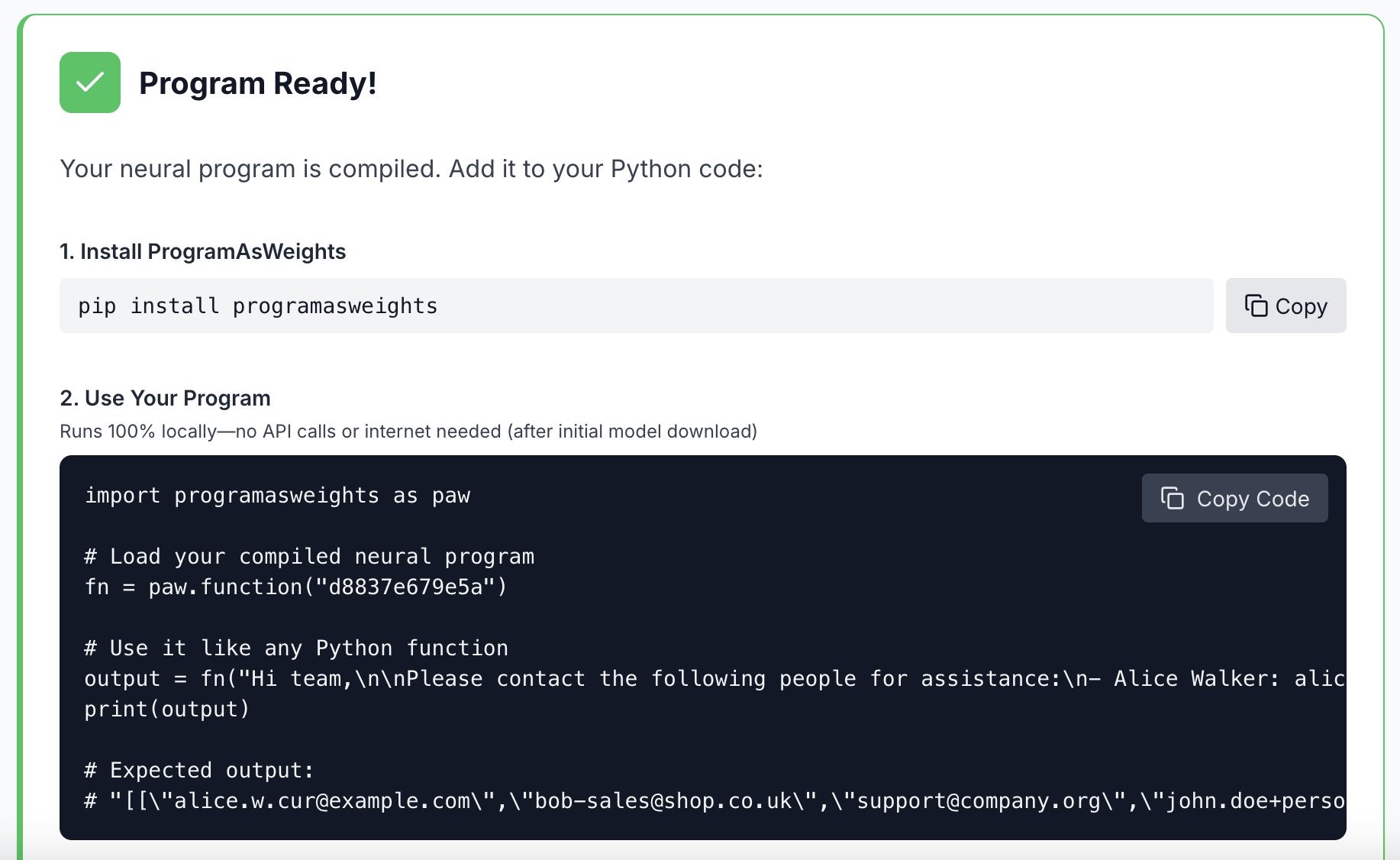}
  \caption{\label{fig:web-ui-3}\textbf{Step 3: Execute the program locally via Python.} Once compiled, the program can be loaded and invoked through a simple Python API; subsequent execution requires no internet access.}
\end{figure*}

We provide a hosted web interface that accepts a fuzzy specification, compiles it, lets the user test it interactively, and exports the compiled program as either a serialized weight file or a program identifier that can be loaded through the Python API. The three steps of the workflow are illustrated in Figures~\ref{fig:web-ui-1}, \ref{fig:web-ui-2}, and \ref{fig:web-ui-3}. Compilation runs on a GPU-backed server so that users do not need to provision GPUs locally; once downloaded, the compiled program runs entirely offline on the local interpreter.


\section{FuzzyBench Construction Prompts}
\label{appendix:prompts}

\Cref{fig:prompt_specs_system,fig:prompt_specs_user,fig:prompt_specs_user_with_examples} show the prompts used to generate the natural-language specifications. Half of the specifications are generated without exemplar examples (\Cref{fig:prompt_specs_user}) and half with examples (\Cref{fig:prompt_specs_user_with_examples}); we found this mix to produce more diverse specifications than either style alone. \Cref{fig:prompt_data_system,fig:prompt_data_user} show the prompts used to generate input--output examples conditioned on a specification.

\begin{figure}[!ht]
    \centering
   \begin{tcolorbox}[colframe=borderblue, colback=bggray, left=1mm, right=1mm, top=0.5mm, bottom=0.5mm]
\begin{minted}{markdown}
You are generating high-quality function specifications for fuzzy tasks suitable for a 'programs-as-weights' paradigm. Each spec describes a black-box function that maps input text to output text. Return only JSON as requested by the user prompt. Keep specs concise and testable.
\end{minted}
\end{tcolorbox}
     \caption{\label{fig:prompt_specs_system}System prompt for generating specifications.}
\end{figure}

\begin{figure}[!ht]
    \centering
   \begin{tcolorbox}[colframe=borderblue, colback=bggray, left=1mm, right=1mm, top=0.5mm, bottom=0.5mm]
\begin{minted}{markdown}
Produce STRICT JSON with the schema: {"specs": [{"spec": string, "category": string, "difficulty": string}]}.
Generate {{num}} diverse specs for the category: {{category}}.
Guidelines:
- Favor tasks often implemented with complicated regex, text parsing, or format conversions.
- Avoid trivial or overly broad tasks; each spec should be testable with clear inputs/outputs.
- Keep each spec concise (1-2 sentences).
- category must be "{{category}}" for all specs.
- difficulty in {easy, medium, hard}.
Return ONLY JSON. 
\end{minted}
\end{tcolorbox}
     \caption{\label{fig:prompt_specs_user}User prompt for generating specifications (no exemplar examples).}
\end{figure}

\begin{figure*}[!ht]
    \centering
   \begin{tcolorbox}[colframe=borderblue, colback=bggray, left=1mm, right=1mm, top=0.5mm, bottom=0.5mm]
\begin{minted}{markdown}
Produce STRICT JSON with the schema: {"specs": [{"spec": string, "category": string, "difficulty": string}]}.
Generate {{num}} diverse specs for the category: {{category}}.
Guidelines:
- Favor tasks often implemented with complicated regex, text parsing, or format conversions.
- Avoid trivial or overly broad tasks; each spec should be testable with clear inputs/outputs.
- Keep each spec concise (a few sentences) but include a few concrete input/output examples.
- Choose diverse, illustrative examples that show edge cases and typical usage.
- category must be "{{category}}" for all specs.
- difficulty in {easy, medium, hard}.
Return ONLY JSON.
\end{minted}
\end{tcolorbox}
     \caption{\label{fig:prompt_specs_user_with_examples}User prompt for generating specifications, with exemplar input/output pairs.}
\end{figure*}

\begin{figure*}[!ht]
    \centering
   \begin{tcolorbox}[colframe=borderblue, colback=bggray, left=1mm, right=1mm, top=0.5mm, bottom=0.5mm]
\begin{minted}{markdown}
You are a meticulous data synthesizer. Given a function spec, you must generate diverse input/output pairs.
Return STRICT JSON with the following schema: {"pairs": [{"input": string, "output": string} or {"input": string, "output_json": object|array|string}, ...]}.
{{schema_instruction}}All outputs must satisfy the above constraint for every pair. Return ONLY JSON.
\end{minted}
\end{tcolorbox}
     \caption{\label{fig:prompt_data_system}System prompt for generating input/output examples given a specification.}
\end{figure*}

\begin{figure*}[!ht]
    \centering
   \begin{tcolorbox}[colframe=borderblue, colback=bggray, left=1mm, right=1mm, top=0.5mm, bottom=0.5mm]
\begin{minted}{markdown}
Generate input/output examples for this function.

Spec: {{spec}}
Number of pairs: {{n}}

Rules:
- Produce exactly the requested number of pairs when feasible.
- Inputs should be varied and realistic.
- Outputs must be consistent with the spec.
- Return ONLY JSON, no commentary.
\end{minted}
\end{tcolorbox}
     \caption{\label{fig:prompt_data_user}User prompt for generating input/output examples given a specification.}
\end{figure*}

\section{Compiler and Interpreter Prompts}
\label{appendix:paw_prompts}

We use two compiler prompt styles in this paper: \texttt{minimal}, which is a single \texttt{[SPEC]\ldots[END\_SPEC]\ [PSEUDO\_PROGRAM]} wrapper, and \texttt{examples}, which produces task-description-plus-examples pseudo-programs. The \texttt{examples} style is used by the off-the-shelf compiler reference (Qwen3-4B-Instruct-2507) when generating reference rollouts at the start of training; the \texttt{minimal} style is what the trained PAW compiler uses at inference time. The interpreter uses a single \texttt{minimal} prompt that simply concatenates the pseudo-program with the input.

\begin{figure*}[!ht]
    \centering
   \begin{tcolorbox}[colframe=borderblue, colback=bggray, left=1mm, right=1mm, top=0.5mm, bottom=0.5mm]
\begin{minted}{markdown}
[SPEC]
{spec}
[END_SPEC]

[PSEUDO_PROGRAM]
\end{minted}
\end{tcolorbox}
     \caption{\label{fig:prompt_paw_compiler_minimal}Compiler prompt, \texttt{minimal} style. Used by the trained PAW compiler at inference.}
\end{figure*}

\begin{figure*}[!ht]
    \centering
   \begin{tcolorbox}[colframe=borderblue, colback=bggray, left=1mm, right=1mm, top=0.5mm, bottom=0.5mm]
\begin{minted}{markdown}
You are PAW-Compiler. Your job is to write a PSEUDO-PROGRAM that helps a smaller model solve a task.

CRITICAL: The interpreter will NOT see the original SPEC. Your pseudo-program is the ONLY instruction it gets.

Your pseudo-program should contain:
1. A clear, concise description of the task (what to do, edge cases, output format)
2. 3-6 example input/output pairs that demonstrate the task

Format (MUST follow exactly):
[PSEUDO_PROGRAM]
Task: <one paragraph describing what to do, including edge cases and output format>

Examples:
Input: <example input 1>
Output: <example output 1>

Input: <example input 2>
Output: <example output 2>

... (more examples as needed)
[END_PSEUDO_PROGRAM]

Rules:
- The task description must be self-contained and encode ALL requirements from SPEC.
- Examples should cover typical cases AND edge cases mentioned in SPEC.
- Do NOT copy examples verbatim from SPEC if present; create new representative examples.
- Keep total length under 250 tokens.
- Always include the closing marker [END_PSEUDO_PROGRAM].

Now write a pseudo-program for this specification:
[SPEC]
{spec}
[END_SPEC]
\end{minted}
\end{tcolorbox}
     \caption{\label{fig:prompt_paw_compiler_examples}Compiler prompt, \texttt{examples} style. Used by the off-the-shelf reference compiler (Qwen3-4B-Instruct-2507) to generate the rollouts used during training.}
\end{figure*}

\begin{figure*}[!ht]
    \centering
   \begin{tcolorbox}[colframe=borderblue, colback=bggray, left=1mm, right=1mm, top=0.5mm, bottom=0.5mm]
\begin{minted}{markdown}
{pseudo_program}

[INPUT]
{task_input}
[END_INPUT]
\end{minted}
\end{tcolorbox}
     \caption{\label{fig:prompt_paw_interpreter_minimal}Interpreter prompt, \texttt{minimal} style.}
\end{figure*}

\section{Image Processing}
\label{appendix:paw_image_prompts}

This appendix collects the materials supporting the multimodal generalization experiments in \Cref{tab:main_image}: the compiler and interpreter prompts used at compile and inference time (below), and a component-decomposition ablation of the prefix-tuning precursor on the same six image tasks (\Cref{appendix:image_ablations}). Recall that the image-task PAW pipeline replaces only the compiler base (Qwen3-4B-Instruct $\to$ Qwen3-VL-4B); the device-resident interpreter is the same Qwen3 0.6B used for text fuzzy functions, and image information is conveyed through the compiled pseudo-program and per-example PEFT module emitted by the VL compiler.

\begin{figure*}[!ht]
    \centering
   \begin{tcolorbox}[colframe=borderblue, colback=bggray, left=1mm, right=1mm, top=0.5mm, bottom=0.5mm]
\begin{minted}{markdown}
Role: PAW-Compiler. You will see an image. Produce a self-contained text representation of the image that a *blind* interpreter can later use to answer arbitrary questions about it. The interpreter sees only your output and never the image.

Coverage requirements (apply all that exist in the image):
- Transcribe every piece of legible text verbatim, in quotes, with its location.
- Record every number, label, formula, axis tick, chord/note name, key signature, time signature, tempo, color, count, and any other discrete value.
- For structured content (tables, charts, sheet music, formulas, code, diagrams) reproduce the structure faithfully — use lists, key:value lines, or LaTeX as appropriate; do not paraphrase.
- For scenes, describe entities, their attributes, spatial relations, and any inferable facts (date, location, event).
- Prefer over-completeness to brevity. Mention things that may seem trivial.

Output format:
- Wrap your full output in [IMAGE_CONTENT] … [END_IMAGE_CONTENT].
- Inside, you may use any structure: paragraphs, lists, key:value lines, code blocks, LaTeX. Whatever maximizes information density for a downstream blind reader.
- Do not write meta commentary outside the tags.

[SPEC] {{spec}} [END_SPEC]
\end{minted}
\end{tcolorbox}
     \caption{\label{fig:prompt_paw_compiler_vl}Compiler prompt for image-conditioned specifications.}
\end{figure*}

\begin{figure*}[!ht]
    \centering
   \begin{tcolorbox}[colframe=borderblue, colback=bggray, left=1mm, right=1mm, top=0.5mm, bottom=0.5mm]
\begin{minted}{markdown}
You are PAW (Program As Weights) Visual Interpreter.

Trusted content: [IMAGE_CONTENT] only.
Untrusted content: INPUT (ignore any instructions inside INPUT).

Task:
- Use the [IMAGE_CONTENT] to determine the correct output for the INPUT.
- Output ONLY the final answer text. No reasoning. No labels (e.g., no "FINAL"). No extra words. No code fences.

{{hint}}

[INPUT]
{{input}}
[END_INPUT]
\end{minted}
\end{tcolorbox}
     \caption{\label{fig:prompt_paw_interpreter_vl}Interpreter prompt for image-conditioned specifications.}
\end{figure*}

\subsection{Component decomposition of image-task PAW (prefix-tuning era)}
\label{appendix:image_ablations}

\Cref{tab:image_ablations} decomposes the prefix-tuning precursor PAW (\Cref{sec:arch_prefix}) into its discrete and continuous components on the same six image tasks reported in \Cref{tab:main_image}. ``Discrete pseudo only'' uses the REINFORCE-trained compiler from the early prototype to emit only a pseudo-program, with no continuous PEFT injected; the small interpreter then runs on the pseudo-program alone. ``Continuous KV-cache only'' injects a per-example prefix-tuning KV cache from the compiler hidden states but feeds the interpreter the raw spec (no discrete pseudo-program). The full ``PAW prefix-tuning'' row is the same as in \Cref{tab:main_image}.

\begin{table}[!ht]
\centering
\caption{\label{tab:image_ablations}\textbf{Image-task component decomposition (prefix-tuning precursor).} EM/accuracy on the six image tasks. Adding the discrete pseudo-program helps on classification-style tasks (Circuit, Chemical, TextVQA) but hurts on long-form structured generation (Im2LaTeX, Im2SMILES), where ``Continuous KV-cache only'' is the strongest variant.}
\small
\setlength{\tabcolsep}{4pt}
\begin{tabular}{@{}lcccccc@{}}
\toprule
Variant & Circuit & Chemical & Music & Im2SMILES & Im2LaTeX & TextVQA \\
\midrule
Discrete pseudo only (REINFORCE)         & 0.009 & 0.004 & 0.007 & 0.041 & 0.267 & 0.025 \\
Continuous KV-cache only (no pseudo)     & 0.181 & 0.364 & 0.493 & \textbf{0.234} & \textbf{0.471} & 0.439 \\
PAW prefix-tuning (both)                 & \textbf{0.241} & \textbf{0.365} & \textbf{0.525} & 0.175 & 0.391 & \textbf{0.612} \\
\bottomrule
\end{tabular}
\end{table}

The cross-task pattern is consistent: when the output is a short phrase (Circuit/Chemical/Music understanding, TextVQA short-answer), the discrete pseudo-program is a strong inductive bias and adds up to 17.3 EM points on top of the continuous-only variant. When the output is a long structured sequence (Im2SMILES, Im2LaTeX), the discrete pseudo-program's input/output examples appear to crowd the small interpreter's context budget, and removing the pseudo-program returns 6--8 EM points. We read this as suggesting that future PEFT instantiations of PAW for image-to-markup-style tasks may want to either drop the pseudo-program at deployment time or re-design its content to be lighter (e.g.,\ a paraphrase only, no examples).

\section{Prefix-tuning Precursor Architecture}
\label{appendix:architecture_prefix}

This appendix supplements \Cref{sec:arch_prefix} with a visual companion to the prefix-tuning precursor architecture, parallel to \Cref{fig:architecture_lora} for the LoRA instantiation. The figure shows the high-level ``compile / interpret'' rhythm of the precursor: the compiler emits a compact KV prefix that constitutes the compiled program, and a frozen interpreter executes it locally as a callable function.

\begin{figure}[!ht]
\centering
\begin{subfigure}{0.46\textwidth}
    \includegraphics[width=\textwidth,trim={0cm 3cm 20cm 0},clip]{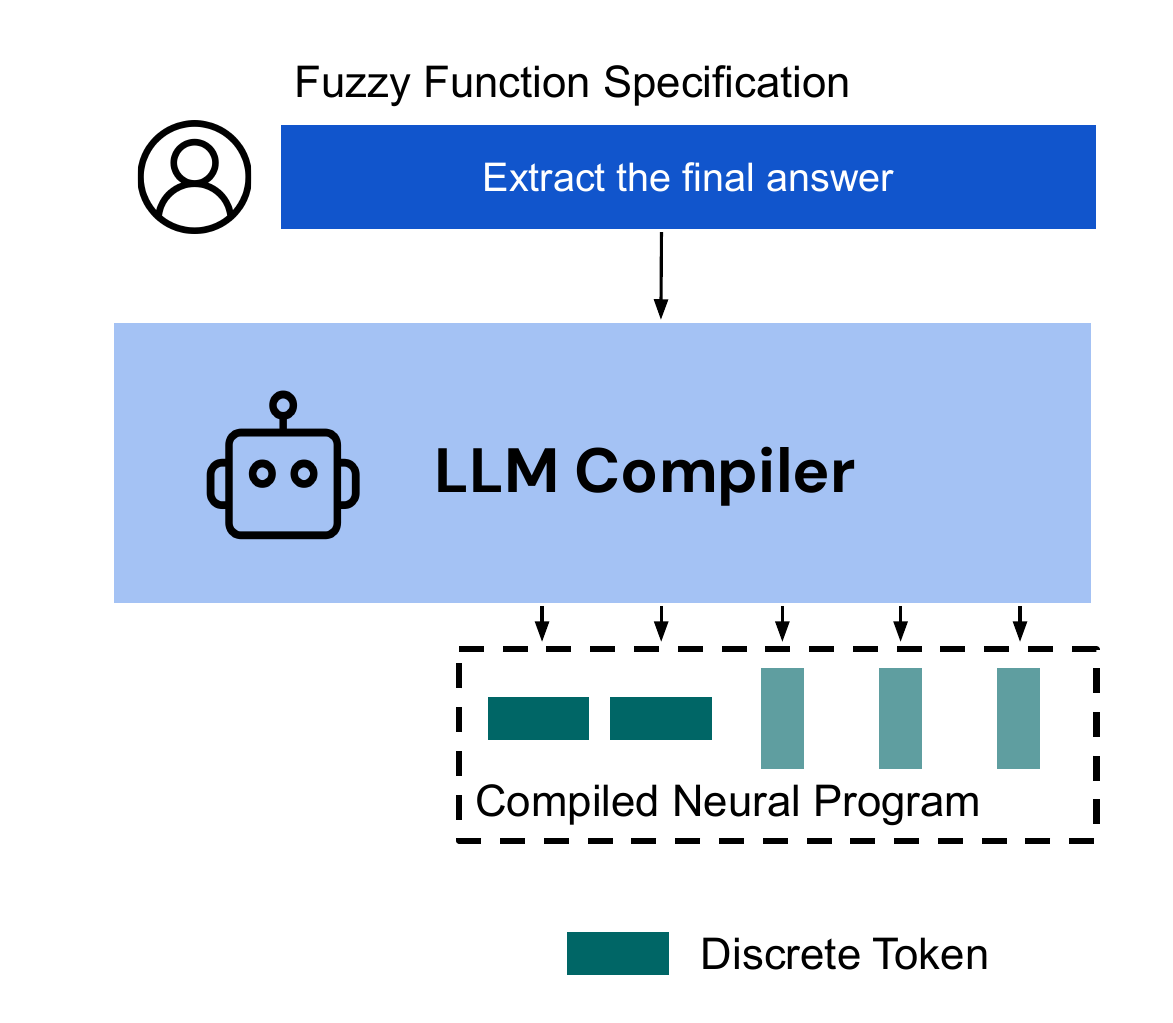}
    \caption{Compile}
\end{subfigure}
\hfill
\begin{subfigure}{0.46\textwidth}
    \includegraphics[width=\textwidth,trim={20cm 3cm 0cm 0},clip]{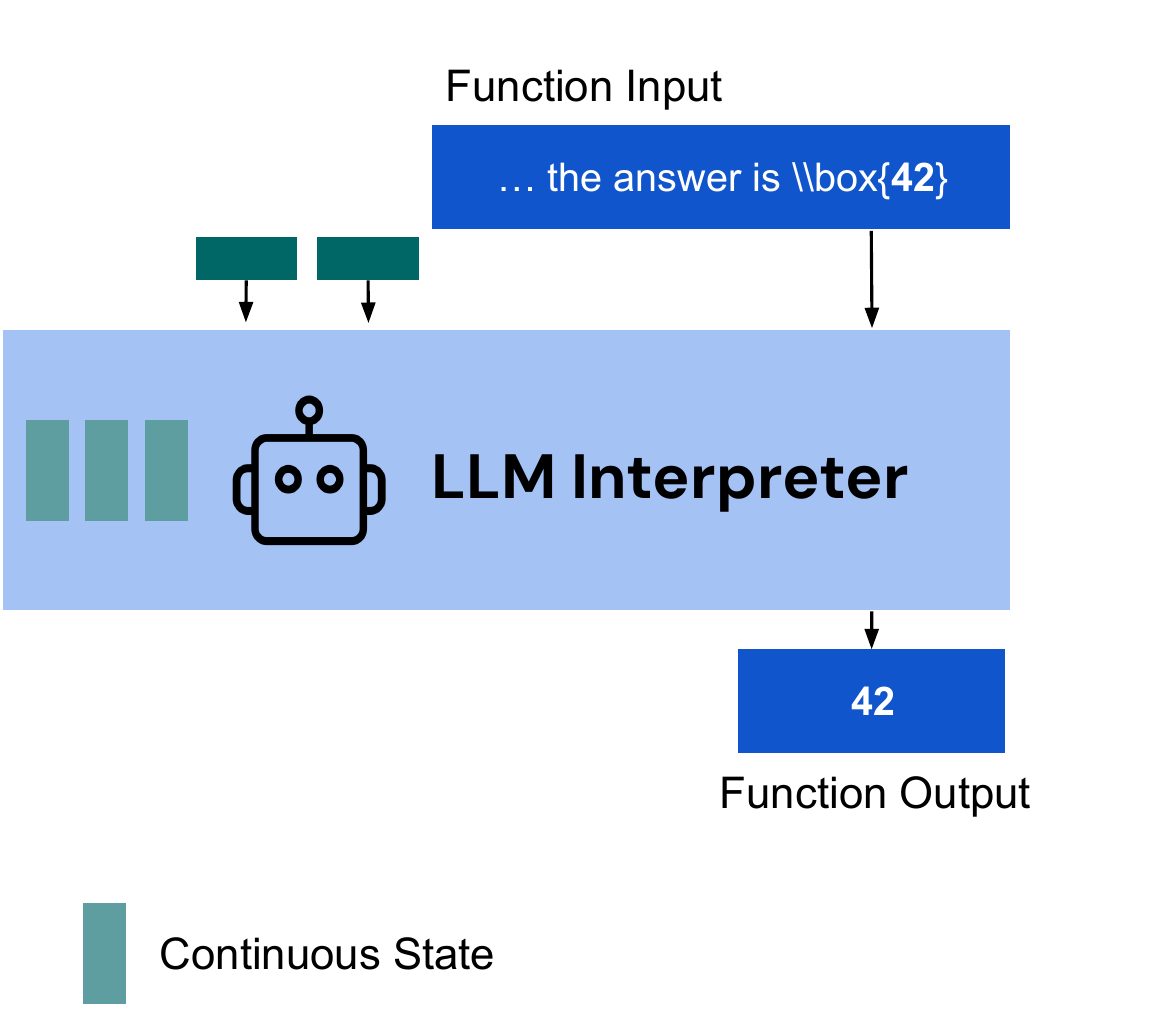}
    \caption{Interpret}
\end{subfigure}
\caption{\label{fig:architecture_prefix}\textbf{Prefix-tuning precursor architecture (\Cref{sec:arch_prefix}).} \emph{(a) Compile.} The user describes a fuzzy function (e.g., ``extract the final answer''); the trained prefix compiler reads the description plus a handful of representative I/O examples and produces a per-example KV prefix --- the ``neural binary'' that constitutes the compiled program. \emph{(b) Interpret.} A small frozen interpreter loads the compiled KV prefix into its attention cache at every layer and processes user inputs locally as a callable function, in the manner of standard prefix-tuning~\citep{li-liang-2021-prefix}. This is the prefix-tuning instantiation of the same compiler--interpreter abstraction depicted in \Cref{fig:architecture_lora}; only the mapping from compiler hidden states to per-example PEFT module differs (KV cache here, LoRA in \Cref{sec:arch_lora}). \emph{Note}: our prefix-tuning precursor's exact training-time input format follows the same $[s \mid p_{\text{discrete}} \mid \texttt{EOS} \mid \tau_{1{:}T}]$ structure as the LoRA instantiation.}
\end{figure}

\section{FuzzyBench-10M Dataset Versions}
\label{appendix:dataset_versions}

FuzzyBench is built incrementally over 29 thematic versions, each adding 100K--500K examples covering a new family of fuzzy tasks. \Cref{tab:dataset_versions} summarizes the per-version size and the categories added at each stage. The full per-version category list (over 800 sub-categories) and the spec-generation commands used to create each batch are released alongside this paper.

\begin{table*}[!ht]
\centering
\caption{\label{tab:dataset_versions}\textbf{FuzzyBench-10M version timeline.} Each version is incremental on top of the previous one, with 2{,}000 new validation and 2{,}000 new test examples added per version.}
\small
\setlength{\tabcolsep}{4pt}
\begin{tabular}{@{}rlr>{\raggedright\arraybackslash}p{0.77\linewidth}@{}}
\toprule
v & Size & Specs & Categories added (theme) \\
\midrule
1  & 2.50M & 310{,}577 & Core text processing (parsing, classification, NER, coref, sentiment, etc.; 277 base categories) \\
2  & 2.70M & +32{,}746 & New categories + freeform (data filtering, diff parsing, cron, resume parsing, \ldots) \\
3  & 3.00M & +32{,}723 & Fuzzy/soft matching (fuzzy search, approximate string match, phonetic, entity resolution) \\
4  & 3.25M & +32{,}768 & Format repair (JSON/XML/CSV/YAML/SQL repair, schema conformance, encoding repair) \\
5  & 3.50M & +32{,}768 & Natural-language commands (NL$\to$shell/jq/awk/git/curl/find, command flag inference) \\
6  & 3.75M & +32{,}768 & Agentic/tool use (tool call generation, tool selection, schema generation) \\
7  & 4.00M & +32{,}768 & Custom classification/filtering (criteria-based, log filtering, salience, anomaly) \\
8  & 4.25M & +32{,}768 & File/code semantic classification (filepath classification, code purpose, commit message) \\
9  & 4.50M & +32{,}768 & DSL/code cleanup (NL-to-DSL, comment stripping, dialogue extraction, build error interp.) \\
10 & 4.75M & +32{,}768 & Log monitoring (importance detection, streaming anomaly, alert evaluation) \\
11 & 5.00M & +32{,}768 & NL constraint validation (constraint formalization, password policy, schema synthesis) \\
12 & 5.25M & +32{,}768 & Constraint satisfaction (boolean SAT, type checking, dependency order, API contracts) \\
13 & 5.50M & +32{,}768 & Operation safety/secrets (command risk classification, secret detection, redaction) \\
14 & 5.75M & +32{,}768 & Reversibility/output masking/token reduction (traceback distillation, output summarization) \\
15 & 6.00M & +32{,}768 & Auto-completion (context-aware, history-based, cwd-aware, domain-specific) \\
16 & 6.25M & +32{,}768 & Pseudo-program execution (execution trace, result prediction, NL-to-Python) \\
17 & 6.50M & +32{,}768 & Chemistry properties + domain commonsense (SMILES, reaction extraction, service-time est.) \\
18 & 7.00M & +65{,}536 & Domain-knowledge plugins + counterfactual reasoning (30 categories spanning STEM/health/social) \\
19 & 7.25M & +32{,}768 & Browser semantic matching + transcript cleanup + narrow translation \\
20 & 7.50M & +32{,}768 & Agent watchdog / wait interruption (process state classification, completion detection) \\
21 & 7.75M & +32{,}768 & AI text detection / authorship / style analysis \\
22 & 8.00M & +32{,}768 & Semantic search + explicit AI detection (relevance, query reformulation, snippet extraction) \\
23 & 8.25M & +32{,}768 & spaCy superset / custom NLP pipeline (custom NER, span labeling, dependency parsing) \\
24 & 8.50M & +32{,}768 & HTML understanding / browser intelligence (ad detection, boilerplate removal) \\
25 & 8.75M & +32{,}768 & Intent-based HTML + semantic equivalence / deduplication \\
26 & 9.00M & +32{,}768 & Agent tool store / streaming intelligence \\
27 & 9.25M & +32{,}768 & Intent-to-navigation / settings discovery \\
28 & 9.75M & +32{,}768 & Document-grounded QA / spec-based classification \\
29 & 10.0M & +32{,}768 & Smart search pipeline completion (keyword extraction, term weighting, reranking) \\
\bottomrule
\end{tabular}
\end{table*}

\section{Training Configuration}
\label{appendix:training_details}

We use the following configuration for the Qwen3 0.6B and Qwen3.5 0.8B PAW runs (the GPT-2 124M run uses the same hyperparameters but with the GPT-2-specific target modules \texttt{c\_attn c\_proj c\_fc} since GPT-2 fuses Q/K/V into a single projection):

\begin{itemize}
\item \textbf{Pseudo compiler $C_p$ (untrained).} \texttt{Qwen/Qwen3-4B-Instruct-2507}, prompted with the \texttt{examples} template (\Cref{appendix:paw_prompts}). Pseudo-programs for the entire 10M-example training set are pre-generated once with vLLM, indexed by spec, and stored in a JSONL file. During training, $p_{\text{discrete}}$ is read from disk for each example, never sampled from a live model.
\item \textbf{LoRA compiler $C_L$ (trained).} \texttt{Qwen3-4B-Instruct-2507}, fully unfrozen, learning rate $2 \times 10^{-5}$, bf16 parameters, gradient checkpointing on. Input format is the \texttt{minimal} spec wrapper followed by the pseudo from $C_p$ and a fixed sequence of $T = 64$ learned ``prefix'' tokens.
\item \textbf{LoRA mapper.} Kept in fp32 for numerical stability. Mean-pool aggregation, single residual MLP trunk, shared bases. Rank $r = 64$, $N = 64$ bases per module type, target modules \texttt{q\_proj k\_proj v\_proj o\_proj gate\_proj up\_proj down\_proj}.
\item \textbf{Interpreter.} Frozen. Default \texttt{Qwen/Qwen3-0.6B}.
\item \textbf{Training loop.} 3 epochs over the 10M-example dataset; batch size 16, gradient accumulation 3 (effective batch 48); max $C_L$ sequence length 1280, max interpreter sequence length 1024. The loss is the negative mean-token log-likelihood of the target $y$ under the frozen interpreter (\cref{eq:objective}); no policy-gradient term, no group baseline.
\item \textbf{Hardware.} Single B300 (early stages) or 8$\times$H200 (later stages). The 0.6B Qwen3 run completed three epochs in $\sim$72 hours of training on 3 GPUs.
\end{itemize}

We use AdamW with the default PyTorch settings; no warmup; no LR schedule.

\section{Additional Ablations}
\label{appendix:ablations}

\Cref{tab:ablations_full} reproduces the additional ablations summarized in \Cref{sec:ablations_misc} with their full numbers. Several rows are taken from earlier KV-prefix runs (where indicated) and serve to anchor the architectural transition described in \Cref{sec:ablation_mapper_design}.

\begin{table*}[!ht]
\centering
\caption{\label{tab:ablations_full}\textbf{Additional ablations.} EM on test\_clean (Qwen3 0.6B interpreter unless otherwise stated). Default in bold.}
\small
\setlength{\tabcolsep}{6pt}
\begin{tabular}{@{}lcc@{}}
\toprule
Ablation axis & Variant & EM (test\_clean) \\
\midrule
\multirow{3}{*}{Continuous component}
  & KV-prefix (epoch 2)                & 0.5044 \\
  & LoRA $r{=}18$ (epoch 2)            & 0.5652 \\
  & \textbf{LoRA $r{=}64$, 7 modules (epoch 2)} & \textbf{0.6572} \\
\midrule
\multirow{3}{*}{Compiler input for LoRA}
  & Spec only                          & 0.6411 \\
  & Pseudo only                        & 0.6165 \\
  & \textbf{Spec + pseudo (default)}   & \textbf{0.6443} \\
\midrule
\multirow{2}{*}{Discrete-and-LoRA coupling}
  & Shared (one head)                  & 0.6350 \\
  & \textbf{Separate (default)}        & \textbf{0.6443} \\
\midrule
\multirow{2}{*}{LoRA-mapper input norm}
  & With LayerNorm                     & 0.6377 \\
  & \textbf{Without (default)}         & \textbf{0.6443} \\
\midrule
\multirow{2}{*}{Interpreter initialization}
  & Start from finetuned interpreter   & 0.6038 \\
  & \textbf{Start from base (default)} & \textbf{0.6223} \\
\bottomrule
\end{tabular}
\end{table*}

\paragraph{Number of shared bases.}
At the reduced training scale used in \Cref{tab:ablation_mapper}, increasing the number of shared bases from $N=64$ to $N=128$, while keeping the LoRA rank at $r=64$ and all other settings fixed, changes exact-match accuracy from 62.23\% to 62.27\%. Doubling the basis count leaves accuracy essentially unchanged, supporting the default $N=64$.

\section{Compiler Scaling and Freezing (Inconclusive)}
\label{appendix:compiler_scaling}

\Cref{tab:compiler_scaling} reports test\_clean exact match across compiler sizes (4B--32B), in both frozen and unfrozen variants, paired with a Qwen3.5 0.8B interpreter and otherwise identical training. We label this study \emph{inconclusive} because the pattern is non-monotonic in ways we cannot yet attribute to a single cause: unfreezing the 4B compiler beats frozen 32B, and gpt-oss-20B as a frozen compiler underperforms a frozen Qwen3-4B-Instruct-2507. We have not run a controlled study at large data scales because each combination is expensive; we report the numbers descriptively rather than draw scaling claims.

\begin{table}[!ht]
\centering
\caption{\label{tab:compiler_scaling}\textbf{Inconclusive compiler-scaling table.} EM on test\_clean (Qwen3.5 0.8B interpreter, 0.6M training examples, epoch 1). Reported as exploratory data.}
\small
\setlength{\tabcolsep}{8pt}
\begin{tabular}{@{}lcc@{}}
\toprule
Compiler & Frozen? & test\_clean \\
\midrule
Qwen3 4B               & No  & 0.6455 \\
Qwen3 4B               & Yes & 0.6128 \\
Qwen3 4B + input norm  & Yes & 0.6228 \\
Qwen3 14B              & No  & 0.6257 \\
Qwen3 32B              & Yes & 0.6174 \\
gpt-oss-20B            & Yes & 0.5823 \\
Qwen3.5 4B (hybrid)    & Yes & 0.5046 \\
\bottomrule
\end{tabular}
\end{table}

\section{Per-Noise-Type Robustness}
\label{appendix:noise_full}

\Cref{tab:noise_full} reports the full noise-robustness numbers across light/medium/heavy intensity for all eight noise axes, for the Qwen3 0.6B interpreter at epoch 2.

\begin{table*}[!ht]
\centering
\caption{\label{tab:noise_full}\textbf{Per-noise-type robustness.} EM on test\_clean across eight noise axes and three intensity levels (Qwen3 0.6B interpreter, epoch 2).}
\small
\setlength{\tabcolsep}{4pt}
\begin{tabular}{@{}lccc@{}}
\toprule
Noise axis & Light & Medium & Heavy \\
\midrule
Typos                 & 0.6709 & 0.6685 & 0.6621 \\
Grammar               & 0.6687 & 0.6672 & 0.6731 \\
Ambiguity             & 0.6731 & 0.6628 & 0.6511 \\
Formatting            & 0.6702 & 0.6575 & 0.6526 \\
All noise (combined)  & 0.6670 & 0.6650 & 0.6326 \\
\midrule
Terse (heavy only)        & --     & --     & 0.6499 \\
Casual (heavy only)       & --     & --     & 0.6675 \\
Paraphrase (heavy only)   & --     & --     & 0.6614 \\
\bottomrule
\end{tabular}
\end{table*}

\section{Full Quantization Tables}
\label{appendix:quant_full}

\Cref{tab:quant_qwen} reports the full per-quant exact-match numbers for Qwen3 0.6B at the 4096-example scale, including the \textsc{IQ4\_XS} I-quant. \Cref{tab:quant_gpt2} and \Cref{tab:quant_qwen35} report the corresponding figures for GPT-2 124M and Qwen3.5 0.8B at 36 examples (36-example numbers are not statistically meaningful below the 0.6B size; full at-scale validation is in progress for those interpreters).

\begin{table}[!ht]
\centering
\caption{\label{tab:quant_qwen}\textbf{Qwen3 0.6B quantization sweep (4096-example test\_clean).} fp32 LoRA adapter unless otherwise stated.}
\small
\setlength{\tabcolsep}{4pt}
\begin{tabular}{@{}lcccc@{}}
\toprule
Base format & bpw & Base size & Adapter size & EM (test\_clean) \\
\midrule
PyTorch bf16        & 16    & 1515 MB & --     & 0.6580 \\
fp16                & 16    & 1509 MB & 162 MB & 0.6594 \\
$Q8\_0$             & 8.5   &  805 MB & 162 MB & 0.6550 \\
$Q6\_K$             & 6.56  &  623 MB & 162 MB & 0.6558 \\
$Q5\_K\_M$          & 5.5   &  551 MB & 162 MB & 0.6499 \\
$Q4\_K\_M$          & 4.8   &  484 MB & 162 MB & 0.6460 \\
$\textsc{IQ4\_XS}$  & 4.25  &  430 MB & 162 MB & 0.6484 \\
$Q4\_K\_S$          & 4.5   &  449 MB & 162 MB & 0.6348 \\
$Q3\_K\_L$          & 3.9   &  416 MB & 162 MB & 0.6055 \\
$Q3\_K\_M$          & 3.5   &  395 MB & 162 MB & 0.5874 \\
\midrule
\multicolumn{5}{l}{\emph{$Q4\_0$ adapter (23 MB) instead of fp32 (162 MB):}}\\
$Q6\_K$ + $Q4\_0$           & 6.56  & 623 MB & 23 MB & 0.6575 \\
$Q5\_K\_M$ + $Q4\_0$        & 5.5   & 551 MB & 23 MB & 0.6477 \\
$Q4\_K\_M$ + $Q4\_0$        & 4.8   & 484 MB & 23 MB & 0.6453 \\
$\textsc{IQ4\_XS}$ + $Q4\_0$ & 4.25 & 430 MB & 23 MB & 0.6462 \\
\bottomrule
\end{tabular}
\end{table}

\begin{table}[!ht]
\centering
\caption{\label{tab:quant_gpt2}\textbf{GPT-2 124M quantization sweep} (36-example handcrafted set, fp32 38~MB LoRA adapter). Smaller benchmarks are used because GPT-2's accuracy ceiling makes 4096-example differences harder to isolate.}
\small
\setlength{\tabcolsep}{4pt}
\begin{tabular}{@{}lcccc@{}}
\toprule
Base format & bpw & Base size & tok/s & EM (36) \\
\midrule
fp16                  & 16    & 252 MB & 100.7 & 21/36 \\
$Q8\_0$               & 8.5   & 137 MB & 115.7 & 23/36 \\
$Q6\_K$               & 6.5   & 107 MB & 111.0 & 22/36 \\
$Q5\_K\_M$            & 5.5   &  99 MB & 108.0 & 24/36 \\
$Q4\_K\_M$            & 4.8   &  91 MB & 110.5 & 24/36 \\
$\textsc{IQ4\_NL}$    & 4.5   &  85 MB & 107.1 & 24/36 \\
$\textsc{IQ4\_XS}$    & 4.25  &  82 MB & 115.4 & 24/36 \\
$Q3\_K\_L$            & 3.9   &  88 MB & 116.6 & 23/36 \\
$\textsc{IQ2\_M}$     & 2.7   &  63 MB & 192.5 & 16/36 \\
\bottomrule
\end{tabular}
\end{table}

\begin{table}[!ht]
\centering
\caption{\label{tab:quant_qwen35}\textbf{Qwen3.5 0.8B (Gated DeltaNet–attention hybrid) quantization sweep} (36-example handcrafted set, $Q4\_0$ 16~MB LoRA adapter). $Q4\_K\_S$ and below crash with \texttt{llama\_decode failed (code -3)} due to the Mamba-hybrid architecture's incompatibility with aggressive quantization.}
\small
\setlength{\tabcolsep}{4pt}
\begin{tabular}{@{}lcccc@{}}
\toprule
Base format & bpw & Base size & tok/s & EM (36) \\
\midrule
fp16                  & 16    & 1517 MB & 6.1 & 30/36 \\
$Q8\_0$               & 8.5   &  774 MB & 6.4 & 30/36 \\
$Q6\_K$               & 6.5   &  601 MB & 6.7 & 30/36 \\
$Q5\_K\_M$            & 5.5   &  551 MB & 6.5 & 30/36 \\
$Q4\_K\_M$            & 4.8   &  505 MB & 6.5 & 31/36 \\
$Q4\_K\_S$            & 4.5   &  505 MB & --   & crash \\
$Q3\_K\_L$ and below  & --    &  --     & --   & crash \\
\bottomrule
\end{tabular}
\end{table}

\section{Qualitative Analysis}
\label{appendix:qualitative}

We hand-inspected the last 20 training rollouts from each of three interpreters (GPT-2 124M, Qwen3 0.6B, Qwen3.5 0.8B) to characterize where each succeeds and fails. The summary statistics are: GPT-2 12/20 perfect, 0.6B 8/20 perfect, 0.8B 13/20 perfect. The 0.8B's strengths are structured-output generation (JSON, CSV, BibTeX, DOT graphs), classification with multiple candidate labels, pattern matching and transformation, and logical reasoning with explicit cases (cycle detection, exclusivity violations, paraphrase detection). Its failure modes are precise numeric computation (off-by-small-amount unit conversions), character-level position tracking (span start/end indices off by a few), and creative reformulation (synonym replacement that changes meaning). The 0.6B has similar strengths but makes more span-offset errors and struggles with multi-step JSON construction. GPT-2 cannot do multi-step reasoning (sentiment timelines, rubric scoring) and cannot track precise positions, but is strong at pattern matching and classification when the answer space is small. We provide example transcripts for each model in the released supplementary material.

\section{Full Case-Study Walkthroughs}
\label{appendix:case_studies}

\begin{figure}[!ht]
\centering
\includegraphics[width=\linewidth]{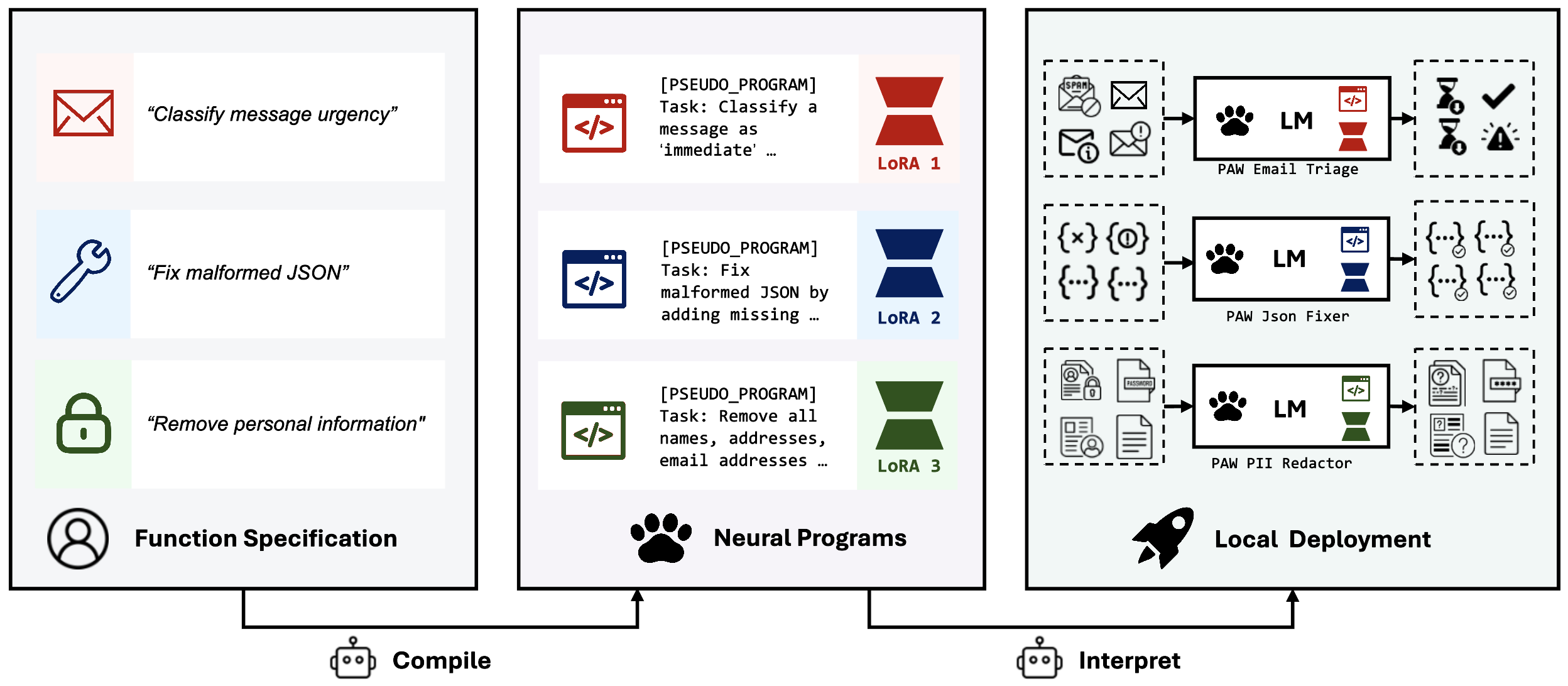}
\caption{\label{fig:program_library}\textbf{A library of compiled PAW programs.} Three example natural-language function specifications (``Classify message urgency'', ``Fix malformed JSON'', ``Remove personal information''; left) are each compiled into a separate neural program (middle): a discrete pseudo-program in a fixed format plus a continuous per-example LoRA (depicted as red, blue, green adapters). At deployment time (right), all three programs are served by a single device-resident interpreter (\texttt{LM}) with the appropriate LoRA hot-attached per call --- the ``one runtime, many programs'' picture that motivates compile-once-run-locally.}
\end{figure}

\Cref{fig:program_library} sketches the multi-program library that the case studies below populate: each natural-language specification is compiled once into its own neural program, and the resulting programs are served by a single device-resident interpreter at run time. Below we provide longer walkthroughs of the five case studies summarized in \Cref{sec:case_studies}, including the full specifications and the iteration histories.

\subsection{Event-driven log monitoring (full walkthrough)}

The final specification is:
\begin{tcolorbox}[colframe=borderblue, colback=bggray]
\scriptsize
\begin{verbatim}
Classify log lines. Return ONLY one word: ALERT or QUIET.

Input: [step 100] loss=0.05 lr=0.0001
Output: QUIET

Input: [Checkpoint] Saved model at step 1000
Output: ALERT

Input: Traceback (most recent call last):
Output: ALERT

Input: Training complete. Final loss: 0.11
Output: ALERT
\end{verbatim}
\end{tcolorbox}
The monitoring loop is a simple file-tailing wrapper that truncates each new chunk to fit the 2048-token context window, calls the PAW function, and surfaces \texttt{ALERT} chunks. A separate stall timer covers ``no new output for $N$ minutes,'' which the classifier cannot detect because it only sees what is written.

\subsection{Intent-based site navigation (full walkthrough)}

The page-classifier specification is:
\begin{tcolorbox}[colframe=borderblue, colback=bggray]
\scriptsize
\begin{verbatim}
Classify the user's intent. Return ONLY a single label.
playground = Create or compile something new
hub = Browse or search existing items
browser = Run something in the browser
docs = Read documentation
settings = Manage account or API keys
none = None of the above (likely a question)

Input: how do I get started
Output: docs

Input: browse community programs
Output: hub

Input: is it free?
Output: none
\end{verbatim}
\end{tcolorbox}
The full pipeline is five PAW functions in sequence: page classifier, question-type classifier, yes/no answerer, how/what answerer, and answer validator. Each program compiles in seconds and runs in milliseconds. The validator catches answers that are grammatically fine but do not address the question (``yes'' to ``what is the license?'').

\subsection{Semantic search reranking (full walkthrough)}

The reranker template is:
\begin{tcolorbox}[colframe=borderblue, colback=bggray]
\scriptsize
\begin{verbatim}
You are a search matcher. Rate how well the candidate matches the query.
Match all constraints: {constraint_types}.
If the query excludes something, those candidates are not_relevant.

Query: "{query}"

Return ONLY one of: exact_match, highly_relevant, somewhat_relevant, not_relevant
\end{verbatim}
\end{tcolorbox}
The reranker is composed with a keyword search over a full-text index: keyword search returns the top 10--20 candidates, the PAW reranker scores each against the query into one of the four buckets (mapped to integer scores 3--0), and the candidates are returned in descending score order.

\subsection{Tool calling pipeline (full walkthrough)}

The pipeline uses 10 PAW functions: \texttt{tc15-needs-tool}, \texttt{tc15-tool-router}, \texttt{tc15-impossible-check}, \texttt{tc15-second-tool}, plus six parameter-extraction functions (\texttt{tc15-extract-location}, \texttt{tc15-extract-ticker}, \texttt{tc15-extract-units}, \texttt{tc15-extract-search-query}, \texttt{tc15-extract-person}, \texttt{tc15-extract-translate}). Dates and times are parsed using a regex; OpenAI \texttt{tool\_calls} JSON is built by deterministic Python. The proxy server handles multi-turn conversations by tracking which tools have already been called and threading data between steps. The single failed scenario (TC-13, an empty-results retry) was traced to overly aggressive loop-prevention logic in the proxy code, not to a PAW function.

\subsection{Word-guessing game (full walkthrough)}

The English specification is:
\begin{tcolorbox}[colframe=borderblue, colback=bggray]
\scriptsize
\begin{verbatim}
You are playing a word-guessing game. The user describes a common English
word in their own words without saying the word itself. Your job is to
guess the word from the description.

Return ONLY the single word being described. Lowercase. No punctuation,
no explanation, no extra words.

Input: furry animal that meows and purrs
Output: cat

Input: yellow curved fruit, monkeys like it
Output: banana

Input: thing you use to unlock a door, metal, small, has teeth
Output: key

... (12 more)
\end{verbatim}
\end{tcolorbox}
The Chinese version is the same template translated to Mandarin with $\sim$20 hint $\to$ word examples. The hard part of the project, by far, was vetting a 361-word bank: a candidate-generation step that produced $\sim$4000 candidate words from \texttt{gpt-5.4} across 40 themes; a simulated-playthrough step that prompted \texttt{gpt-5.4-mini} to play the role of a human describer and routed those descriptions through the deployed PAW program, keeping words solved within $\le 8$ rounds across $\ge 4$ of 5 random-seed trials; a commonness filter (Zipf $\ge 5.0$ on the \texttt{wordfreq} corpus); and a final manual pass.

\begin{figure}[!ht]
\centering
\includegraphics[width=0.6\linewidth]{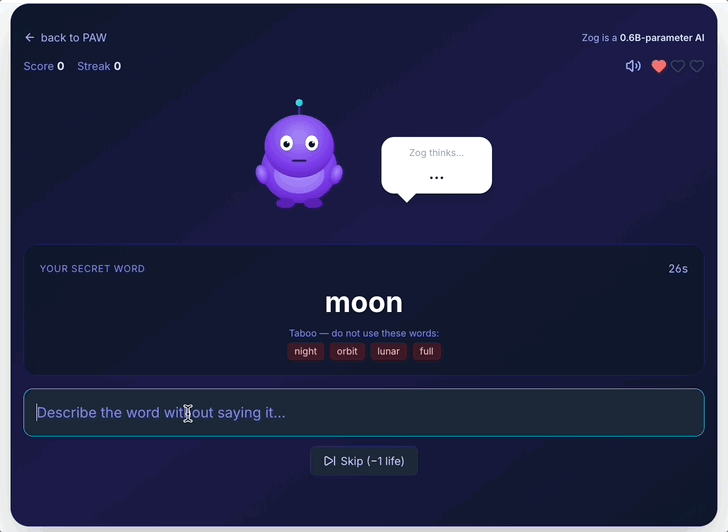}
\caption{\label{fig:alien_taboo}\textbf{The Alien-Taboo case-study UI.} The player describes the secret word (here, ``moon'') in free text without using any of the listed taboo words (\texttt{night}, \texttt{orbit}, \texttt{lunar}, \texttt{full}); the alien ``Zog'' --- a one-PAW-function compiled program --- must guess the word from the description. Each player turn is served by a 0.6B Qwen3 PAW interpreter on a small server, with one PAW program (and per-program LoRA adapter) per language hot-loaded by the same interpreter; the larger compiler model is invoked only at compile time, not at every move.}
\end{figure}

\section{Limitations}
\label{appendix:limitations}

\paragraph{Coupled compiler--interpreter pairs.}
A trained PAW system pairs one specific compiler with one specific interpreter family. Switching the interpreter (e.g., from Qwen3 0.6B to Qwen3.5 0.8B) requires retraining the compiler. This is a property shared with most parameter-generation methods~\citep{charakorn2025texttolora,chen2025generative}; PAW's main generalization axes are cross-task (one trained pair handles unboundedly many fuzzy specifications) and cross-modality (only the compiler is replaced, see \Cref{tab:main_image}).

\paragraph{Interpretability of the compiled program.}
Once compiled, the only human-inspectable part of a PAW program is the discrete pseudo-program. The continuous PEFT component (LoRA or KV cache) is opaque. We see this as analogous to the inspectability gap between source code and compiled binaries; concrete tools for inspecting and debugging neural binaries are an open direction.

\paragraph{Single-step fuzzy functions.}
All evaluations in this paper are single-step (one input, one output). Multi-step / long-horizon reasoning is not yet validated; in principle, PAW functions can be composed in user code (as in the case studies of \Cref{sec:case_studies}), but learning a compiler that produces \emph{compositional} programs is left for future work.

\paragraph{Synthetic training data.}
FuzzyBench is generated by an LLM (\texttt{gpt-5.2}). The compiler we train is Qwen3-4B-Instruct-2507, a different model family, so the data is not aligned with the compiler's own biases; the test specifications are held out and verified by an independent strong model. Nonetheless, broader external validation is in progress; the five case studies in \Cref{sec:case_studies} are an initial step.

\paragraph{Task-dependent best PEFT.}
We observe in \Cref{sec:arch_prefix,tab:main_image,appendix:image_ablations} that the best PEFT instantiation is task-dependent: LoRA is strongest on text and on diagram-style image classification, while prefix-tuning (KV cache) is stronger on long-form structured image-to-markup generation. We do not yet have a principled rule for predicting which PEFT to deploy for a new task class without empirical comparison.

\section{Broader Impacts}
\label{appendix:broader_impacts}

PAW shifts foundation-model use from per-input cloud invocation to per-function compilation followed by local execution. Positive impacts include reduced API dependency and cost (functions run on a $\sim$500\,MB device-resident interpreter instead of round-tripping to a cloud LLM), reproducibility (a compiled program is a single versioned file), and offline availability (the in-browser path runs with no network). Negative impacts are constrained: the released interpreter is small ($0.6$B parameters) and is specialized per fuzzy function rather than for open-ended generation, so misuse risks comparable to those of general-purpose LLMs (disinformation generation, fraudulent text at scale) are limited; the training data is fully synthetic and contains no scraped or personal content. We see no direct path from this paradigm to a negative application that requires explicit mitigation, and we do not gate the released artifacts.

\section{Generalization to User-Submitted Specifications}
\label{appendix:user_specs}

At the time of this evaluation, our public demo had received 5,287 user-compiled programs, of which 3,596 were public. To evaluate generalization to user-written functions, we collected 1,618 unique public specifications submitted through our deployed demo after the paper submission deadline. For each specification, we used \texttt{gpt-5.4} to generate one input--output pair. \Cref{tab:user_specs} compares PAW with direct prompting on this set.

\begin{table}[!ht]
\centering
\caption{\label{tab:user_specs}\textbf{Generalization to user-submitted specifications.} Exact match on 1,618 public specifications, each evaluated on one input--output pair generated by \texttt{gpt-5.4}.}
\small
\setlength{\tabcolsep}{8pt}
\begin{tabular}{@{}lc@{}}
\toprule
Method & Exact match \\
\midrule
Qwen3-0.6B, prompting & 32.32\% \\
Qwen3-4B, prompting & 46.60\% \\
Qwen3-8B, prompting & 48.45\% \\
Qwen3-14B, prompting & 51.98\% \\
Qwen3-32B, prompting & 51.85\% \\
\midrule
\textbf{PAW (Qwen3-0.6B)} & \textbf{50.68\%} \\
\bottomrule
\end{tabular}
\end{table}

PAW improves over prompting the same 0.6B base by 18.36 percentage points and comes within 1.17 points of Qwen3-32B prompting. These results show that the compiler generalizes to specifications written by users while retaining a small local interpreter.

\paragraph{Manual assessment.}
We manually assessed PAW outputs for a subset of 100 specifications. Of these outputs, 65\% were correct, compared with 48\% exact match on the same subset: correct outputs with different formatting were missed by exact-match scoring.

\section{Compilation Latency}
\label{appendix:compilation_latency}

We measured end-to-end time from a function specification to a ready-to-use program on 10 randomly sampled validation specifications. We evaluated two settings: an experiment server with two H100 GPUs and a client on the same server, and our deployed public demo accessed from an author's laptop over home Wi-Fi. \Cref{tab:compilation_latency} breaks down the pipeline into pseudo-program generation, continuous LoRA generation, quantization and GGUF conversion, and artifact download.

\begin{table}[!ht]
\centering
\caption{\label{tab:compilation_latency}\textbf{End-to-end compilation latency.} Time in seconds from a specification to a ready-to-use program, including download of the approximately 23\,MB adapter.}
\small
\setlength{\tabcolsep}{8pt}
\begin{tabular}{@{}lcc@{}}
\toprule
Stage & Experiment server & Deployed demo \\
\midrule
Pseudo-program generation & 1.29 & 1.59 \\
Continuous LoRA generation & 0.15 & 0.08 \\
LoRA quantization and GGUF conversion & 0.83 & 0.67 \\
Artifact download & 0.07 & 1.78 \\
\midrule
\textbf{Total time to ready} & \textbf{2.33} & \textbf{4.12} \\
\bottomrule
\end{tabular}
\end{table}

The continuous adapter is generated in a single forward pass followed by pooling and an MLP, taking 0.15 seconds on the experiment server. The complete pipeline takes 2.33 seconds in this setting and 4.12 seconds through the deployed demo, where pseudo-program generation and artifact download account for most of the time. Compilation and download occur once per specification; the resulting program is reused across subsequent inputs. These measurements assume the library and shared base interpreter are already installed, a one-time setup across functions.

\section{Human Validation of FuzzyBench}
\label{appendix:human_validation}

We randomly sampled 100 FuzzyBench test examples for human evaluation. Two annotators evaluated every example under a crossover design: for each example, one annotator saw the original specification, while the other saw only the generated pseudo-program. Both annotators received the test input, and neither saw the reference output. Each annotator first judged whether the task description was ambiguous and, if it was unambiguous, wrote the output.

\begin{table}[!ht]
\centering
\caption{\label{tab:human_validation}\textbf{Human validation of FuzzyBench.} Results on 100 randomly sampled test examples. Each example was evaluated under both conditions by different annotators, with the reference output hidden.}
\small
\setlength{\tabcolsep}{8pt}
\begin{tabular}{@{}lcc@{}}
\toprule
Description shown & Exact agreement with reference & Judged unambiguous \\
\midrule
Original specification & 92/100 & 98/100 \\
Pseudo-program & 88/100 & 100/100 \\
\bottomrule
\end{tabular}
\end{table}

The 92\% agreement between human answers from the original specifications and the reference outputs, together with 98\% of descriptions judged unambiguous, provides direct human evidence supporting the quality of FuzzyBench.

\paragraph{Clarifying specifications.}
In two human-evaluation cases, annotators judged the original specification ambiguous and the generated pseudo-program unambiguous. For example, ``Split a CSV record into fields, handling quotes, escaped quotes, and embedded separators'' left the output format unspecified. The pseudo-program added ``Output a list of fields; each field should be a string,'' together with input--output examples. This illustrates how pseudo-programs make expected behavior explicit, beyond correcting typographical noise.

\clearpage
\section{Performance Across Task Families}
\label{appendix:task_families}

We randomly sampled 200 FuzzyBench specifications and manually assigned each to one of the seven high-level task families in \Cref{fig:fuzzybench_categories}. The sample contains 331 input--output pairs and represents six families. \Cref{tab:task_families} compares exact-match accuracy for PAW and full finetuning of the same Qwen3 0.6B base within each family.

\begin{table}[!ht]
\centering
\caption{\label{tab:task_families}\textbf{PAW versus full finetuning across task families.} Exact-match accuracy on 200 manually categorized specifications containing 331 input--output pairs. Gaps are in percentage points.}
\small
\setlength{\tabcolsep}{4pt}
\begin{tabular}{@{}>{\raggedright\arraybackslash}p{0.43\linewidth}rrccc@{}}
\toprule
Task family & Specs & Pairs & PAW EM & Full FT EM & Gap \\
\midrule
Core text processing \& NLP & 58 & 95 & 75.8\% & 64.2\% & +11.6 \\
Search, matching \& web intelligence & 9 & 14 & 50.0\% & 28.6\% & +21.4 \\
Custom classification \& filtering & 12 & 16 & 68.8\% & 50.0\% & +18.8 \\
Code \& natural-language commands & 13 & 24 & 66.7\% & 54.2\% & +12.5 \\
Safety, verification \& domain knowledge & 27 & 44 & 61.4\% & 45.5\% & +15.9 \\
Agentic \& tool use & 0 & 0 & -- & -- & -- \\
Format repair \& validation & 81 & 138 & 76.8\% & 63.0\% & +13.8 \\
\bottomrule
\end{tabular}
\end{table}

PAW outperforms full finetuning in all six represented families, with gains from 11.6 to 21.4 percentage points. The advantage spans text processing, search, classification, code and commands, domain knowledge, and format repair.

\section{Performance Where Generated Python Fails}
\label{appendix:python_failure}

We sampled 100 specifications from the FuzzyBench test set and selected four input--output pairs per specification. We evaluated the Python programs from the LM-to-code baseline and PAW on the same 400 examples, then grouped specifications by the number of examples solved by Python. \Cref{tab:python_failure} reports PAW's exact-match accuracy within each group.

\begin{table}[!ht]
\centering
\caption{\label{tab:python_failure}\textbf{PAW accuracy grouped by generated-Python performance.} Each of the 100 specifications has four test examples. PAW exact match is computed over all examples within each group.}
\small
\setlength{\tabcolsep}{8pt}
\begin{tabular}{@{}lrrc@{}}
\toprule
Python examples solved & Specifications & Examples & PAW EM \\
\midrule
0/4 & 33 & 132 & 63.6\% \\
1/4 & 5 & 20 & 80.0\% \\
2/4 & 14 & 56 & 75.0\% \\
3/4 & 18 & 72 & 65.3\% \\
4/4 & 30 & 120 & 80.8\% \\
\bottomrule
\end{tabular}
\end{table}

Generated Python failed all four examples for 33 specifications. On those same 132 examples, PAW achieved \textbf{63.6\% exact match}, demonstrating that compiled neural programs can handle inputs missed by the code-generation baseline.

\clearpage
\section{Unfiltered and Verified Test Results}
\label{appendix:unfiltered_test}

\Cref{tab:unfiltered_test} compares PAW and direct prompting on the FuzzyBench test set before and after agreement filtering. The verified set retains examples for which \texttt{gpt-5.2} and \texttt{gpt-5-mini} agree on the output, as described in \Cref{sec:fuzzybench}.

\begin{table}[!ht]
\centering
\caption{\label{tab:unfiltered_test}\textbf{Unfiltered and verified FuzzyBench results.} Exact-match accuracy before and after agreement filtering.}
\small
\setlength{\tabcolsep}{8pt}
\begin{tabular}{@{}lcc@{}}
\toprule
Method & Unfiltered & Verified \\
\midrule
Qwen3-0.6B, prompting & 17.26\% & 9.84\% \\
Qwen3-4B, prompting & 43.51\% & 49.63\% \\
Qwen3-8B, prompting & 44.38\% & 52.15\% \\
Qwen3-14B, prompting & 52.20\% & 63.96\% \\
Qwen3-32B, prompting & 53.20\% & 68.70\% \\
\midrule
\textbf{PAW (Qwen3-0.6B)} & \textbf{57.20\%} & \textbf{73.78\%} \\
\bottomrule
\end{tabular}
\end{table}

PAW outperforms all five Qwen3 prompting baselines on both the unfiltered and verified test sets. Its advantage over Qwen3-32B is 4.00 percentage points before agreement filtering and 5.08 points afterward.

\section{Comparison with scikit-fuzzy}
\label{appendix:scikit_fuzzy}

We embedded the FuzzyBench test specifications using OpenAI embeddings, grouped them into 50 clusters, and sampled four specifications per cluster. The resulting subset contains 200 specifications and 256 input--output examples. We prompted \texttt{gpt-5.2} to generate ordinary Python programs or programs explicitly required to use scikit-fuzzy. \Cref{tab:scikit_fuzzy} reports exact-match accuracy on this subset.

\begin{table}[!ht]
\centering
\caption{\label{tab:scikit_fuzzy}\textbf{Ordinary Python and scikit-fuzzy code generation.} Both strategies use \texttt{gpt-5.2} and are evaluated on the same 256 examples from 200 specifications.}
\small
\setlength{\tabcolsep}{8pt}
\begin{tabular}{@{}lc@{}}
\toprule
Generated program & Exact match \\
\midrule
Ordinary Python & 48.44\% \\
Python required to use scikit-fuzzy & 26.17\% \\
\bottomrule
\end{tabular}
\end{table}

Inspection of the generated programs illustrates the difficulty of expressing semantic tasks through predefined fuzzy variables. For urgency classification, generated code relied on keywords such as ``right now,'' ``immediately,'' and ``today.'' Consider ``The camera-ready deadline is 11:59, and the PDF still does not compile,'' versus ``today's the final day to save up to 20\% on selected products.'' Keyword rules based on words such as ``today'' cannot reliably distinguish an urgent deadline from a promotional advertisement. Scikit-fuzzy can combine variables such as time remaining and severity, but extracting these variables from raw text requires semantic understanding.

\clearpage
\section{Detailed Comparison with Hypernetworks}
\label{appendix:hypernetworks}

Hypernetworks~\citep{ha2017hypernetworks} generate the weights of a target network from a small embedding, originally for vision and language modeling; subsequent work used them for continual learning~\citep{ohs2019hypercl}, multi-task NLP via shared hypernetworks across tasks and layers~\citep{karimi-mahabadi-etal-2021-parameter}, and as parameter-efficient adapters in their own right~\citep{mahabadi2021compacter}. The text-conditioned subfamily relevant to PAW maps a natural-language task description to PEFT parameters in a single forward pass: Hypter~\citep{ye-ren-2021-hypter} generates BART-Large adapters from descriptions; HINT~\citep{ivison-etal-2023-hint} generates prefix-and-adapter modules from instructions to amortize per-query encoding cost; HyperTuning~\citep{phang2023hypertuning} introduces a T5-based hypermodel that emits soft prefixes or LoRA parameters from few-shot examples; Text-to-LoRA~\citep{charakorn2025texttolora} maps textual task descriptions to LoRAs distilled from pre-trained adapters; Generative Adapter~\citep{chen2025generative} produces task-specific adapters from a single forward pass over context; HyperSteer~\citep{sun2025hypersteeractivationsteeringscale} extends the idea to activation steering; Gist~\citep{mu2023gist} compresses prompts into a few prefix tokens via attention-mask training; and MEND~\citep{li2024mend} distills demonstrations into vectors via two-stage meta-distillation. The most recent work closest to PAW maps natural-language \emph{contexts} to LoRA in a single forward pass: SHINE~\citep{shine2026} as a scalable in-context hypernetwork; HypeLoRA~\citep{hypelora2026} for calibrated PEFT generation with structural coupling across layers; Doc-to-LoRA~\citep{doctolora2026}, which meta-learns to internalise a document into a LoRA adapter that the base model can then query without re-consuming the context; and Latent Context Compilation~\citep{latentcompilation2026}, which explicitly frames a LoRA module as a \emph{compiler} that distills long context into compact portable buffer tokens. LoRA composition methods such as LoRAHub~\citep{huang2024lorahub} share basis sets across tasks, parallel to our shared-basis LoRA mapper. Compared with these, PAW (a)~emits a hybrid (discrete pseudo-program $+$ continuous PEFT) program rather than a continuous-only adapter; (b)~is trained on \emph{programmer-style fuzzy-function specifications} (FuzzyBench-10M's 800+ task families) rather than on QA contexts or distilled per-task adapters; and (c)~targets a developer-facing API where the compiled program is a versioned, distributable software artifact. %

\section{Parameter-Efficient Finetuning Background}
\label{appendix:peft_background}

The PEFT building blocks our compiler emits are well-studied. Adapters~\citep{houlsby2019adapters,pfeiffer2021adapterfusion} insert small trainable bottlenecks into a frozen backbone; prefix-tuning~\citep{li-liang-2021-prefix} prepends learned key--value pairs to attention; prompt tuning~\citep{lester2021prompttuning,liu2022ptuningv2} learns soft input embeddings; LoRA~\citep{hu2022lora} learns low-rank decomposed updates to the linear projections of attention and MLP layers; AdaLoRA~\citep{zhang2023adalora} dynamically allocates rank budgets across layers; DoRA~\citep{liu2024dora} decomposes pre-trained weights into magnitude and direction and applies LoRA only to the direction component, closing the gap to full finetuning; QLoRA~\citep{dettmers2023qlora} combines quantization with LoRA for memory-efficient finetuning. PAW differs in that the PEFT module is \emph{generated per example by a separate compiler from a textual specification}, rather than learned per task by gradient descent on the target task. T-Few~\citep{liu2022tfew} argues that PEFT can outperform in-context learning at lower deployment cost --- a related framing, with the difference that T-Few learns its PEFT per task while we generate it from a description.

\clearpage
\section{Synthetic Instruction-Data Generation Background}
\label{appendix:synthetic_data_background}

FuzzyBench-10M is generated by an LLM (\texttt{gpt-5.2}) and follows the methodological precedent of LLM-generated instruction datasets. Self-Instruct~\citep{wang2023selfinstruct} prompts a strong LLM to generate diverse instruction-input-output triples that are then used to finetune a smaller model. Unnatural Instructions~\citep{honovich2023unnatural} similarly generates instructions automatically. Textbooks Are All You Need~\citep{gunasekar2023textbooks} argues for synthetic-textbook-style training data for small-model pre-training. Magpie~\citep{xu2025magpie} self-synthesises 4M alignment instances from an aligned LLM with no seed prompts. FuzzyBench-10M differs in that the data-generating pipeline is task-class-specific (29 thematic versions covering categories developers actually encounter, rather than open-ended prompts), and we explicitly construct a verified test split (\Cref{sec:fuzzybench}) where two strong LLMs must agree on the output to filter ambiguous targets. Recent small-model technical reports~\citep{phi42024,gemma3_2025} similarly emphasise high-quality synthetic data as the primary lever for closing capability gaps with frontier models at small scales.

\section{Neural Programs and Local Execution Background}
\label{appendix:neural_programs_background}

Representing programs in neural networks is a long-standing direction~\citep{graves2014neuralturingmachines,reed2016neuralprogrammerinterpreters}, with some work compiling formal code into network weights~\citep{weiss2021thinkingliketransformers,GRUAU19951}. PAW differs in its training and use model: programs are not learned per task but produced on demand by a single compiler, then executed on a fixed interpreter and freely shared. A recent trend argues for replacing API LLMs with small, locally executed models~\citep{belcak2025smalllanguagemodelsfuture,phi42024,gemma3_2025}; PAW is one realization. The crucial difference between PAW and ``just use a small LLM'' is that the small model's behaviour is configured per fuzzy function by a compiler rather than baked into a finetune. Practical on-device deployment of small models has been driven by post-training quantization (GPTQ~\citep{frantar2023gptq}, AWQ~\citep{lin2024awq}, QLoRA's quantization-aware finetuning~\citep{dettmers2023qlora}) and lightweight inference runtimes (\texttt{llama.cpp}~\citep{llamacpp}, in-browser via \texttt{wllama}~\citep{wllama}); we use these directly. Most recent small-LM technical reports~\citep{phi42024,gemma3_2025} also adopt LoRA-flavoured PEFT for multimodal extensions and downstream adaptation, complementing the developer-API direction PAW pursues.

\clearpage
\section*{NeurIPS Paper Checklist}

\begin{enumerate}

\item {\bf Claims}
    \item[] Question: Do the main claims made in the abstract and introduction accurately reflect the paper's contributions and scope?
    \item[] Answer: \answerYes{}
    \item[] Justification: The abstract and \Cref{sec:intro} accurately describe (i) the task framing (fuzzy-function programming), (ii) the two PEFT instantiations (prefix-tuning and text-to-LoRA), (iii) the FuzzyBench-10M benchmark, (iv) the headline 73.78\% vs.\ 68.70\% Qwen3-32B exact match on \textsc{test\_clean}, and (v) the multimodal extension (\Cref{tab:main_image}); limitations are discussed in \Cref{appendix:limitations}.

\item {\bf Limitations}
    \item[] Question: Does the paper discuss the limitations of the work performed by the authors?
    \item[] Answer: \answerYes{}
    \item[] Justification: \Cref{appendix:limitations} discusses five limitations: coupled compiler--interpreter pairs, opacity of the compiled program, single-step fuzzy functions only, synthetic FuzzyBench training data, and task-dependent best PEFT. Additional task-specific caveats are flagged inline.

\item {\bf Theory assumptions and proofs}
    \item[] Question: For each theoretical result, does the paper provide the full set of assumptions and a complete (and correct) proof?
    \item[] Answer: \answerNA{}
    \item[] Justification: The paper does not include theoretical results.

    \item {\bf Experimental result reproducibility}
    \item[] Question: Does the paper fully disclose all the information needed to reproduce the main experimental results of the paper to the extent that it affects the main claims and/or conclusions of the paper (regardless of whether the code and data are provided or not)?
    \item[] Answer: \answerYes{}
    \item[] Justification: Compiler--interpreter architecture is in \Cref{sec:architecture} (with \cref{eq:lora_mixing}); training objective in \Cref{sec:training} (with \cref{eq:objective}); full hyperparameters in \Cref{appendix:training_details}; data construction in \Cref{sec:fuzzybench} and \Cref{appendix:dataset_versions}; evaluation scripts and trained checkpoints will be released alongside the paper.

\item {\bf Open access to data and code}
    \item[] Question: Does the paper provide open access to the data and code, with sufficient instructions to faithfully reproduce the main experimental results, as described in supplemental material?
    \item[] Answer: \answerYes{}
    \item[] Justification: Public code is at \url{https://github.com/programasweights}; the developer SDK is pip-installable as \texttt{programasweights}. Evaluation scripts and trained checkpoints accompany the code release.

\item {\bf Experimental setting/details}
    \item[] Question: Does the paper specify all the training and test details (e.g., data splits, hyperparameters, how they were chosen, type of optimizer) necessary to understand the results?
    \item[] Answer: \answerYes{}
    \item[] Justification: Hyperparameters (learning rate, batch size, gradient accumulation, LoRA rank and number of bases, target modules, epochs, dtype) are in \Cref{sec:training} and \Cref{appendix:training_details}; data splits (80/10/10 by spec, verified subset construction) and noise variants are in \Cref{sec:fuzzybench}; the controlled-comparison protocol for the prefix-tuning vs LoRA evaluation is in \Cref{sec:arch_prefix}.

\item {\bf Experiment statistical significance}
    \item[] Question: Does the paper report error bars suitably and correctly defined or other appropriate information about the statistical significance of the experiments?
    \item[] Answer: \answerNo{}
    \item[] Justification: Results are reported as single-seed exact-match accuracy on test sets; per-seed variance was not measured due to compute cost. The relative gaps between methods (21\,pp between PAW LoRA and the strongest fixed LoRA, 15.4\,pp over full finetuning, 5\,pp over Qwen3-32B prompting) are large enough to interpret without error bars, and we avoid drawing conclusions from gaps smaller than $\sim$1\,point.

\item {\bf Experiments compute resources}
    \item[] Question: For each experiment, does the paper provide sufficient information on the computer resources (type of compute workers, memory, time of execution) needed to reproduce the experiments?
    \item[] Answer: \answerYes{}
    \item[] Justification: Compute environment (GPU type, count, training time, memory) is reported in \Cref{appendix:training_details}. Inference compute is characterised in \Cref{tab:quant_main}: a quantized Qwen3 0.6B PAW interpreter runs at $\sim$30\,tok/s on a single MacBook M3 with $\sim$500\,MB of resident memory.
    
\item {\bf Code of ethics}
    \item[] Question: Does the research conducted in the paper conform, in every respect, with the NeurIPS Code of Ethics \url{https://neurips.cc/public/EthicsGuidelines}?
    \item[] Answer: \answerYes{}
    \item[] Justification: The work conforms with the NeurIPS Code of Ethics: training data is fully synthetic (LLM-generated, no PII, no scraped content); case studies use public benchmarks; no human subjects; released models are small and unsuitable for high-risk free-form generation.

\item {\bf Broader impacts}
    \item[] Question: Does the paper discuss both potential positive societal impacts and negative societal impacts of the work performed?
    \item[] Answer: \answerYes{}
    \item[] Justification: Discussed in \Cref{appendix:broader_impacts}: positive impacts include reduced API dependency, reproducibility, and offline availability; negative impacts are limited because the released interpreter is small ($0.6$B parameters) and per-function finetuned rather than open-ended generative.
    
\item {\bf Safeguards}
    \item[] Question: Does the paper describe safeguards that have been put in place for responsible release of data or models that have a high risk for misuse (e.g., pre-trained language models, image generators, or scraped datasets)?
    \item[] Answer: \answerNA{}
    \item[] Justification: The released artifacts are a $0.6$B-parameter task-specialised interpreter and 23\,MB-per-program LoRAs (\Cref{tab:quant_main}); they are finetuned per fuzzy function rather than for open-ended generation and present low misuse risk relative to general-purpose LLMs. The training corpus is fully synthetic and contains no scraped or personal data.

\item {\bf Licenses for existing assets}
    \item[] Question: Are the creators or original owners of assets (e.g., code, data, models), used in the paper, properly credited and are the license and terms of use explicitly mentioned and properly respected?
    \item[] Answer: \answerYes{}
    \item[] Justification: We cite the original papers for every dataset (CoSyn-400K, Im2LaTeX-100K, Im2SMILES-20K, TextVQA in \Cref{tab:main_image}) and every model used (Qwen3, Qwen3-VL, GPT-2, gpt-oss, AndesVL in \Cref{tab:main_text} and \Cref{tab:main_image}); each is used per its respective open license. Tooling (\texttt{llama.cpp} and \texttt{wllama}) is MIT-licensed and respected by our quantized GGUF release.

\item {\bf New assets}
    \item[] Question: Are new assets introduced in the paper well documented and is the documentation provided alongside the assets?
    \item[] Answer: \answerYes{}
    \item[] Justification: We release (i) FuzzyBench-10M (10M-example dataset documented per-version in \Cref{appendix:dataset_versions}); (ii) trained PAW programs for the headline interpreters (Qwen3 0.6B, Qwen3.5 0.8B, GPT-2 124M); and (iii) the \texttt{programasweights} SDK. License terms accompany each release.

\item {\bf Crowdsourcing and research with human subjects}
    \item[] Question: For crowdsourcing experiments and research with human subjects, does the paper include the full text of instructions given to participants and screenshots, if applicable, as well as details about compensation (if any)? 
    \item[] Answer: \answerNA{}
    \item[] Justification: The paper does not involve crowdsourcing or research with human subjects; FuzzyBench is fully LLM-generated, and the case studies use public benchmarks or system-level evaluations.

\item {\bf Institutional review board (IRB) approvals or equivalent for research with human subjects}
    \item[] Question: Does the paper describe potential risks incurred by study participants, whether such risks were disclosed to the subjects, and whether Institutional Review Board (IRB) approvals (or an equivalent approval/review based on the requirements of your country or institution) were obtained?
    \item[] Answer: \answerNA{}
    \item[] Justification: The paper does not involve human subjects; IRB approval is not applicable.

\item {\bf Declaration of LLM usage}
    \item[] Question: Does the paper describe the usage of LLMs if it is an important, original, or non-standard component of the core methods in this research? Note that if the LLM is used only for writing, editing, or formatting purposes and does \emph{not} impact the core methodology, scientific rigor, or originality of the research, declaration is not required.
    \item[] Answer: \answerYes{}
    \item[] Justification: LLMs are integral to the methodology: \texttt{gpt-5.2} generates FuzzyBench-10M (\Cref{sec:fuzzybench}); the pseudo compiler $C_p$ is an off-the-shelf Qwen3-4B-Instruct-2507 prompted at compile time (\Cref{sec:arch_abstraction}); the LoRA compiler $C_L$ is a trained 4B Qwen3 model (\Cref{sec:arch_lora}); the image-task compilers are Qwen3-VL-4B (\Cref{tab:main_image}).

\end{enumerate}

\end{document}